\documentclass[journal]{IEEEtran}
\ifCLASSINFOpdf
   \usepackage[pdftex]{graphicx}
\else
\fi
\usepackage{amssymb}
\usepackage{multirow}
\usepackage{subfigure}
\usepackage{float}
\usepackage{epstopdf}

\usepackage{amsmath}
\usepackage{array}
\begin{document}

%\title{Modality-specific Cross-modal Matching \\ with Recurrent Attention Network}
%MSMRAN: Modality-specific Space Measurement with Recurrent Attention Network for Cross-modal Retrieval
%Modality-specific Measurement with Recurrent Attention Network for Cross-modal Retrieval
\title{Modality-specific Cross-modal Similarity Measurement with Recurrent Attention Network}
%Modality-specific Measurement for Cross-modal Similarity Metric with Recurrent Attention Network
%Modality-specific similarity measurement for cross-modal retrieval with recurrent attention network
%Modality-specific Cross-modal Similarity Measurement with Recurrent Attention Network

\author{Yuxin Peng, Jinwei Qi and Yuxin Yuan
	\thanks{This work was supported by National Natural Science Foundation of China under Grants 61371128 and 61532005.}
	\thanks{The authors are with the Institute of Computer Science and Technology,	Peking University, Beijing 100871, China. Corresponding author: Yuxin Peng	(e-mail: pengyuxin@pku.edu.cn).}
}


\markboth{IEEE TRANSACTIONS ON IMAGE PROCESSING}%
{Shell \MakeLowercase{\textit{et al.}}: Bare Demo of IEEEtran.cls for IEEE Journals}



\maketitle

\begin{abstract}
Nowadays, cross-modal retrieval plays an indispensable role to flexibly find information across different modalities of data. Effectively measuring the similarity between different modalities of data is the key of cross-modal retrieval. 
%Different modalities such as image and text may provide asymmetrical information to describe the same semantics, because some modality-specific characteristics within one modality cannot be exactly aligned with other modalities when performing cross-modal matching. 
Different modalities such as image and text have imbalanced and complementary relationships, which contain unequal amount of information when describing the same semantics. For example, images often contain more details that cannot be demonstrated by textual descriptions and vice versa. 
Existing works based on Deep Neural Network (DNN) mostly construct one common space for different modalities to find the latent alignments between them, which lose their exclusive modality-specific characteristics. % and cannot fully exploit the intrinsic information within each modality.
%Thus, treating different modalities equally to find the latent alignments between them may lose their exclusive modality-specific characteristics, which cannot fully exploit the intrinsic information within each modality.
%which leads to misalignment when performing cross-modal matching.
%Different from existing works based on Deep Neural Network (DNN), which mostly construct one common space for different modalities equally, 
Different from the existing works, we propose modality-specific cross-modal similarity measurement (MCSM) approach by constructing independent semantic space for each modality, which adopts end-to-end framework to directly generate modality-specific cross-modal similarity without explicit common representation. 
%For each semantic space, modality-specific characteristics within one modality are fully exploit, while the data of another modality is projected into it to capture the imbalanced and complementary relationships between different modalities. 
%Specifically, we design recurrent attention network to model the modality-specific characteristics with attention mechanism. An attention based joint embedding loss is further proposed to utilize the learned attention weights for guiding the fine-grained cross-modal correlation learning. 
For each semantic space, modality-specific characteristics within one modality are fully exploited by recurrent attention network, while the data of another modality is projected into this space with attention based joint embedding to utilize the learned attention weights for guiding the fine-grained cross-modal correlation learning, which can capture the imbalanced and complementary relationships between different modalities.
%For each semantic space, we design recurrent attention network for one modality to model the fine-grained context information with attention mechanism, aiming to exploit its intrinsic modality-specific characteristics. An attention based joint embedding loss is proposed to project the data of another modality into the semantic space, which can guide the fine-grained cross-modal correlation learning to utilize the imbalance relationships contained in different modalities. 
Finally, the complementarity between the semantic spaces for different modalities is explored by adaptive fusion of the modality-specific cross-modal similarities to perform cross-modal retrieval. 
Experiments on the widely-used Wikipedia and Pascal Sentence datasets as well as our constructed large-scale XMediaNet dataset verify the effectiveness of our proposed approach, outperforming 9 state-of-the-art methods.
\end{abstract}


% \fi
\IEEEpeerreviewmaketitle

\section{Introduction}

\begin{figure}[!t]
	\centering
	\includegraphics[width=0.485\textwidth]{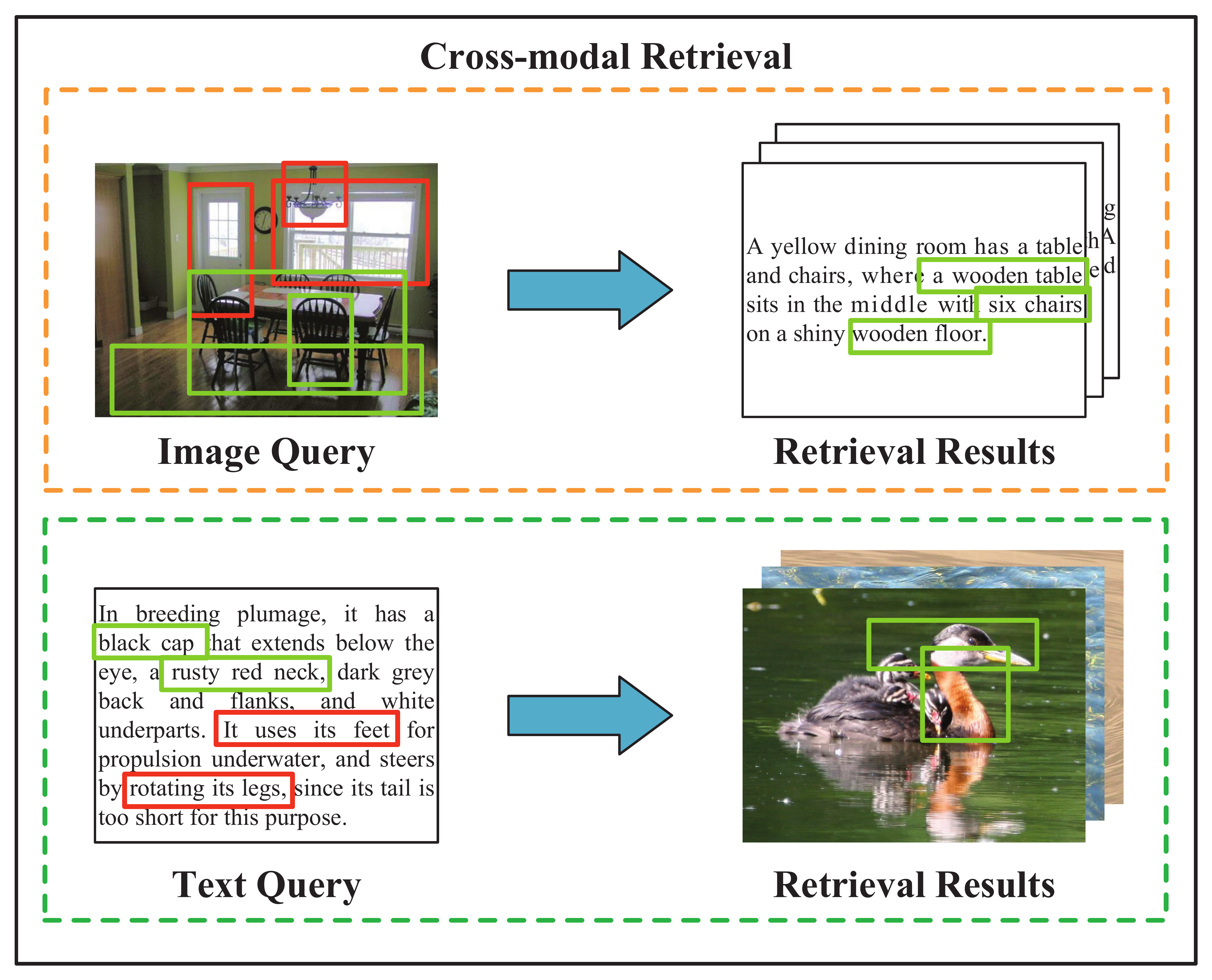}
	\setlength{\abovecaptionskip}{0.3cm}
	\caption{An example of cross-modal retrieval with image and text, which can present retrieval results with different modalities. We can also see that the instances of different modalities have imbalanced and complementary relationships, where the green boxes indicate the fine-grained information described by both image and text, while the red boxes mean extra information contained in only one modality that is not demonstrated by another modality.
		%And green boxes indicate the appropriate alignment between visual and textual fine-grained information, while the red boxes means the misalignment in each modality.
	}
	\label{fig_cross_media}
\end{figure}

\IEEEPARstart{N}{owadays}, multimodal data such as image, video, text and audio, has been widely available on the Internet, which is the fundamental component for promoting artificial intelligence to understand the real world. Some works have been done for breaking the boundaries between different modalities. Cross-modal retrieval has become a highlighted research topic to perform retrieval across different modalities of data, which has great demands with many practical applications, such as search engine and digital libraries. An example of cross-modal retrieval is shown in Figure \ref{fig_cross_media}. 
Different from the traditional single-modal retrieval, such as image retrieval \cite{DBLP:journals/tip/TangLWZ15} and video retrieval \cite{PengCSVT06ClipRetrieval}, which is limited in providing retrieval results of the same single modality with query, cross-modal retrieval is more flexible and useful to retrieve relevant multimodal information by submitting one query of any modality \cite{peng2017overview}. 

The main challenge of cross-modal retrieval is to deal with the inconsistency between different modalities and learn the intrinsic correlation between them. For the fact that different modalities of data has diverse representations and distributions that usually span different feature spaces, this heterogeneous characteristic makes it hard to directly measure the similarity between different modalities, such as an image and an audio clip. To address this issue, some approaches \cite{RasiwasiaMM10SemanticCCA,DBLP:journals/tip/XuSYSL17,DBLP:journals/tip/WangGWHY16,DBLP:journals/tip/HeZWJY15} have been proposed to project the features of different modalities into one common space to learn the common representation, by which the cross-modal similarity can be calculated to perform retrieval between different modalities. Traditional methods build the common space by learning mapping matrices to maximize the correlations of variables from different modalities, such as methods based on Canonical Correlation Analysis (CCA) \cite{RasiwasiaMM10SemanticCCA,DBLP:journals/ijcv/GongKIL14}, graph regularization \cite{ZhaiTCSVT2014JRL,DBLP:journals/pami/WangHWWT16} and learning to rank \cite{DBLP:journals/tip/WuJLTLZZ15,DBLP:journals/tip/WuLSYZRZ16}. Recently, the dramatic advances in deep learning have inspired researchers to bridge the gap between different modalities with Deep Neural Network (DNN). Such methods like \cite{DBLP:conf/ijcai/PengHQ16,feng12014cross,DBLP:conf/cvpr/YanM15} attempt to exploit the advantages of DNN in modeling nonlinear correlation to learn common representation with multilayer network.

\begin{figure}[t]
	\centering
	\begin{minipage}[c]{1.0\linewidth}
		\centering
		\includegraphics[width=1.0\textwidth]{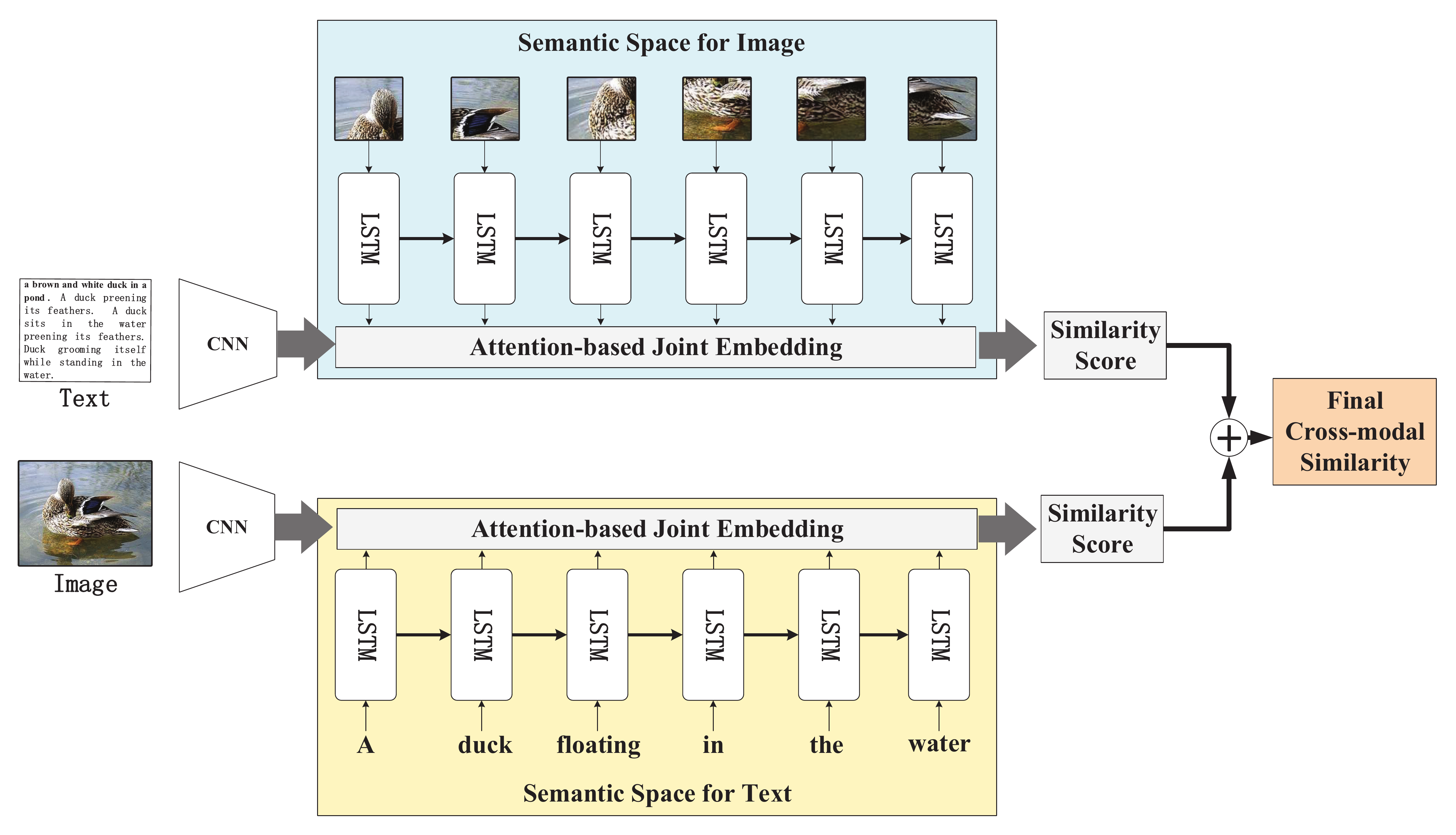}
	\end{minipage}%
	\setlength{\abovecaptionskip}{0.55cm}
	\caption{An overview of the proposed approach, which constructs independent semantic spaces for different modalities, and directly generates modality-specific cross-modal similarities through end-to-end framework without explicit common representation.}
	\label{fig:overview}
\end{figure}

The aforementioned methods mostly project the data of different modalities from their own feature spaces into one single common space equally to find the latent alignments between them, which means \textit{equal} amount of information is captured from the data of different modalities during cross-modal correlation learning.
%The aforementioned methods mostly rely on one common space, where the data of different modalities are treated equally.
Generally speaking, different modalities such as image and text have imbalanced and complementary relationships that provide \textit{unequal} amount of information in describing the same semantics, because some modality-specific characteristics within one modality cannot be exactly aligned with other modalities. % when performing cross-modal correlation learning. 
For example, an image often co-occurs with its corresponding text descriptions on a web page to describe the same semantics such as objects or events. %But the amount of information provided by image and text may be imbalance, 
But the relationships between image and text instance are imbalanced, where not all fine-grained image details can be exactly aligned to the textual descriptions and vice versa. 
Furthermore, image often contains more details which cannot be demonstrated by textual descriptions, that is what we usually say ``A picture is worth a thousand words''. While in other cases, the opposite would happen that textual descriptions contain more semantic information than image when describing some high-level semantics, such as historical events or literature works, where the textual description contains more background information that cannot be shown by its corresponding image.
%Furthermore, in some cases, textual descriptions contain more extra semantic information than image, such as when describing some historical event. While in other cases, the opposite would happen that image contains more details that cannot be demonstrated by textual description. 
As shown in Figure \ref{fig_cross_media}, green boxes indicate the appropriate alignment between visual and textual fine-grained information, while the red boxes mean the misalignment. Therefore, treating different modalities equally to find fine-grained alignments for constructing one common space loses such exclusive and useful modality-specific characteristics, which cannot fully exploit the intrinsic information within each modality.

For addressing the above issues, we aim to preserve the modality-specific characteristics by fully exploiting the fine-grained information within each modality when learning the cross-modal correlation. Thus, modality-specific measurement is required for each modality instead of only constructing one common space. 
Recently, attention mechanism has made great advances in DNN, which allows models to concentrate on the necessary fine-grained parts of visual or textual inputs, and has been successfully applied to various multimodal tasks, such as image caption \cite{DBLP:conf/icml/XuBKCCSZB15} and visual question answering \cite{DBLP:conf/nips/LuYBP16}. Inspired by these advances, we attempt to apply attention mechanism to fully exploit the intrinsic modality-specific fine-grained information for cross-modal retrieval.. 

In this paper, we propose modality-specific cross-modal similarity measurement (MCSM) approach, which constructs independent semantic spaces for different modalities, where the modality-specific characteristics can be fully explored with attention mechanism during cross-modal correlation learning. The modality-specific cross-modal similarity is directly generated from each semantic space through the end-to-end framework without explicit common representation. Figure \ref{fig:overview} shows an overview of our proposed approach. The main contributions of this paper are presented as follows:
\begin{itemize}
	\item{
		%We propose \textit{modality-specific cross-modal similarity measurement} approach to construct independent semantic space for each modality. In each semantic space, the modality-specific characteristics are fully exploited by modeling the fine-grained information within one modality, while the data of another modality is projected into this semantic space to capture the imbalanced and complementary relationships between different modalities during cross-modal correlation learning. 
		\textit{Modality-specific cross-modal similarity measurement.} We construct independent semantic space for each modality. In each semantic space, the modality-specific characteristics are fully exploited by modeling the fine-grained information within one modality, while the data of another modality is projected into this semantic space to capture the imbalanced and complementary relationships between different modalities during cross-modal correlation learning. 
	} 
	\item{
		%We design \textit{recurrent attention network} in each semantic space to capture the modality-specific characteristics including both fine-grained local and context information by recurrent network with attention mechanism. While an \textit{attention based joint embedding loss} is proposed to utilize the learned attention weights for guiding the fine-grained cross-modal correlation learning. %to utilize the imbalance information contained in different modalities.
		\textit{Recurrent attention network with joint embedding.} We design recurrent attention network in each semantic space to capture the modality-specific characteristics including both fine-grained local and context information by recurrent network with attention mechanism. While an attention based joint embedding loss is proposed to utilize the learned attention weights for guiding the fine-grained cross-modal correlation learning.
	} 
	\item {
		%We adopt an \textit{end-to-end} framework in each semantic space to directly generate modality-specific cross-modal similarity, which integrates both representation learning and similarity measurement stages to benefit each other. Furthermore, \textit{adaptive fusion} is adopted to obtain the final similarity for performing cross-modal retrieval, which can fully explore the complementarity between the semantic spaces for different modalities.
		\textit{End-to-end frameworks with adaptive fusion.} We adopt an end-to-end framework in each semantic space to directly generate modality-specific cross-modal similarity, which integrates both representation learning and similarity measurement stages to benefit each other. Furthermore, adaptive fusion is proposed to obtain the final similarity for performing cross-modal retrieval, which can fully explore the complementarity between the semantic spaces for different modalities.
	}
\end{itemize}
Experiments on the widely-used Wikipedia and Pascal Sentence datasets as well as our constructed large-scale XMediaNet dataset show that our proposed MCSM approach outperforms 9 state-of-the-art methods, which can verify the effectiveness of MCSM approach.

The remainder of this paper is organized as follows. We briefly review the related works in Section II. In Section III, our proposed MCSM approach is presented in detail. Then Section IV reports the experimental results as well as analyses. Finally, Section V concludes this paper.

\section{Related Works}

In this section, we briefly review the representative cross-modal retrieval methods with common space learning, as well as recent advances of attention mechanism in DNN.

\subsection{Common Space Learning for Cross-modal Retrieval}

\begin{figure}[!t]
	\centering
	\includegraphics[width=0.487\textwidth]{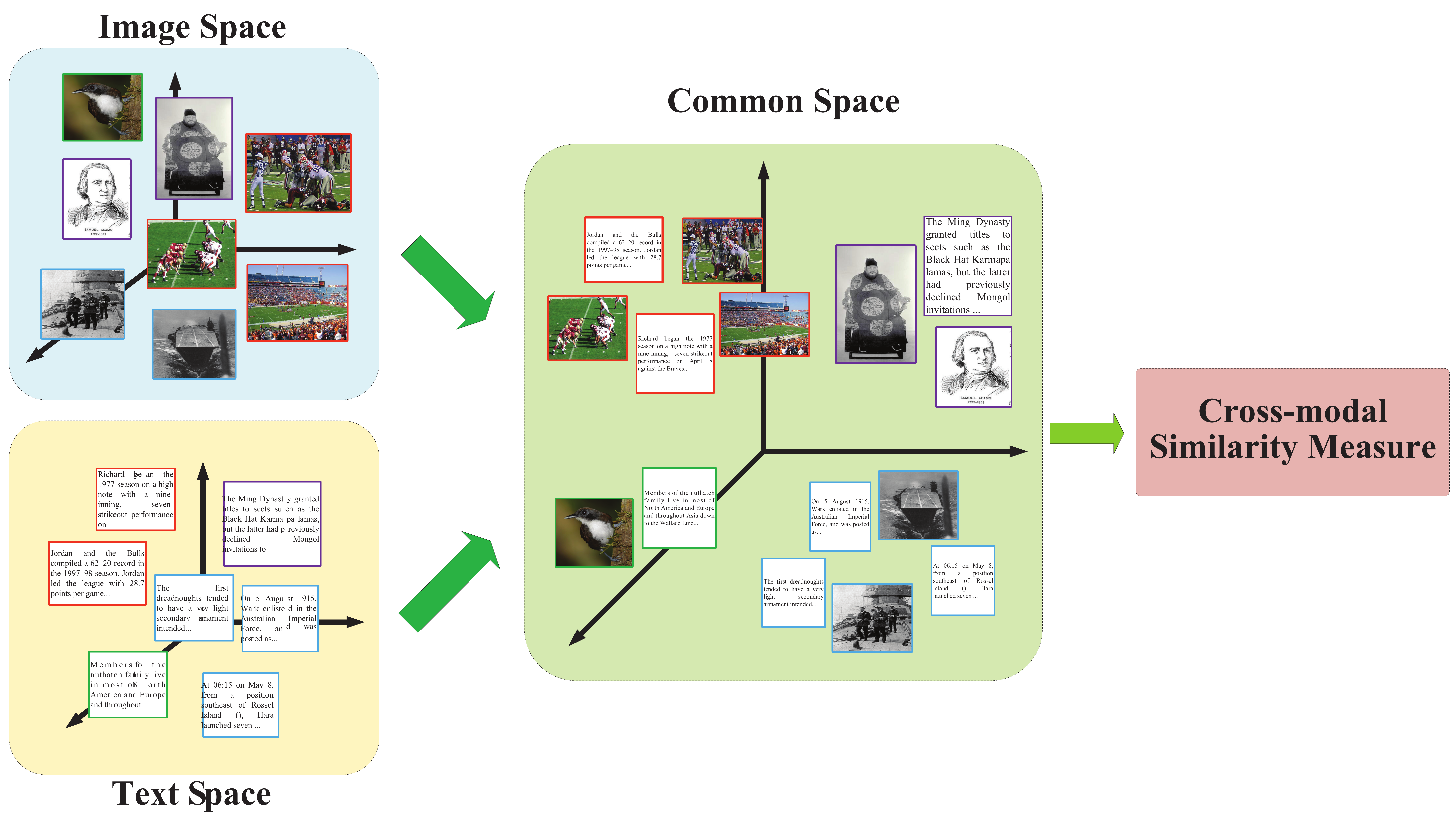}
	\setlength{\abovecaptionskip}{0.0cm}
	\caption{Illustration of mainstream framework of the most cross-modal retrieval methods, which attempts to represent data of different modalities with the same ``feature'' type (namely common representation) in one common space, so that the similarity measure can be directly performed.
	}
	
	\label{fig_CommonSpace}
	
\end{figure}

The current mainstream of cross-modal retrieval methods is to learn an intermediate common space for the features of different modalities, then the cross-modal similarity can be directly measured in one common space. Figure \ref{fig_CommonSpace} illustrates the main framework of such common space learning methods. As indicated in \cite{peng2017overview}, we mainly introduce three categories of existing methods as follows, namely traditional statistical correlation analysis methods, cross-modal graph regularization methods and DNN-based methods.

\subsubsection{Traditional statistical correlation analysis methods}

As the foundation of common space learning methods, these methods mainly optimize the statistical values to learn linear projection matrices, which project features of different modalities into the common space and obtain the common representation. Canonical Correlation Analysis (CCA) \cite{HotelingBiometrika36RelationBetweenTwoVariates}, as one of the most representative works, is a natural solution to maximize the pairwise correlation between the data of different modalities such as image/text pairs \cite{RasiwasiaMM10SemanticCCA}. Furthermore, semantic information is incorporated to extend CCA for improving the accuracy of cross-modal retrieval. For example, Costa et al. \cite{DBLP:journals/pami/PereiraCDRLLV14} integrate semantic category labels to improve the performance of CCA. Multi-view CCA \cite{DBLP:journals/ijcv/GongKIL14} is proposed to construct a third view for modeling the high-level semantics. While Ranjan et al. \cite{DBLP:conf/iccv/RanjanRJ15} propose multi-label CCA, which considers the high-level semantic information in the form of multi-label annotations. Besides, Cross-modal Factor Analysis (CFA) \cite{LiMM03CFA}, as 
one of the alternative methods, is proposed to minimize the Frobenius norm between the data of different modalities after projecting them into one common space.

\subsubsection{Cross-modal graph regularization methods}

Graph regularization \cite{DBLP:conf/colt/BelkinMN04} is widely used to construct a partially labeled graph for semi-supervised learning, which aims to enrich the training set and smooth the solution. Zhai et al. \cite{ZhaiAAAI2013JGRHML} are the first to integrate graph regularization into cross-modal retrieval and propose Joint Graph Regularized Heterogeneous Metric Learning (JGRHML), which constructs the joint graph regularization term in the learned metric space. Furthermore, Joint Representation Learning (JRL) \cite{ZhaiTCSVT2014JRL} is proposed to construct several separate graphs for different modalities and learn projection matrices with the joint consideration of correlation and semantic information. Peng et al. \cite{DBLP:journals/tcsv/PengZZH16} further improve the previous works \cite{ZhaiTCSVT2014JRL,ZhaiAAAI2013JGRHML} by constructing a unified hypergraph to learn the common space for up to five modalities, which also utilizes the fine-grained information at the same time. Besides, Wang et al. \cite{DBLP:journals/pami/WangHWWT16} also adopt multimodal graph regularization term to preserve inter-modality and intra-modality similarity relationships.

\subsubsection{DNN-based methods}

Deep learning has made great advance in multimodal applications, such as image/video classification \cite{DBLP:conf/nips/KrizhevskySH12,DBLP:conf/mm/WuJWYX16} and object recognition \cite{DBLP:conf/cvpr/HeZRS16}. Researches also adopt DNN to perform common space learning to take the advantage of its considerable ability on modeling highly nonlinear correlation. Most of the DNN-based methods construct two subnetworks for different modalities such as image and text, which are linked at joint layer as the common space to model cross-modal correlation. Bimodal Autoencoders (Bimodal AE) \cite{ngiam32011multimodal} is proposed as an extension of Restricted Boltzmann Machine (RBM) to model multiple modalities by minimizing the reconstruction error. 
Srivastava et al. \cite{srivastava2012learning} propose Multimodal Deep Belief Network (Multimodal DBN), which adopts two kinds of DBN for different modalities to model the distribution over their original features, while a joint RBM is adopted on the top of them to model the joint distribution and get the common representation. 
Correspondence Autoencoder (Corr-AE) \cite{feng12014cross} and Deep Canonical Correlation Analysis (DCCA) \cite{DBLP:conf/icml/AndrewABL13} also consist of two subnetworks, while Corr-AE jointly models correlation and reconstruction information, and DCCA combines DNN with CCA to maximize the correlation on the top of two subnetworks. 
Besides, Peng et al. \cite{DBLP:conf/ijcai/PengHQ16} propose Cross-media Multiple Deep Networks (CMDN) to model the intra-modality and inter-modality correlation in a two-stage learning framework. The above works mainly take hand-crafted features as image inputs. Furthermore, Wei et al. \cite{DBLP:journals/tcyb/WeiZLWLZY17} propose deep-SM to perform deep semantic matching with Convolutional Neural Network (CNN), which exploits the strong representation ability of CNN features to improve the retrieval accuracy. He et al. \cite{DBLP:journals/tmm/HeXKWP16} adopt two convolution-based networks to model the matched and mismatched image/text pairs via deep and bidirectional representation learning.

The aforementioned methods mostly construct one common space for different modalities to find the latent alignments between them, which lose exclusive modality-specific characteristics. Therefore, we attempt to adopt modality-specific measurement to fully exploit such modality-specific characteristics by modeling the intrinsic fine-grained information within each modality.

\begin{figure*}[t]
	\centering
	\begin{minipage}[c]{1.0\linewidth}
		\centering
		\includegraphics[width=1.0\textwidth]{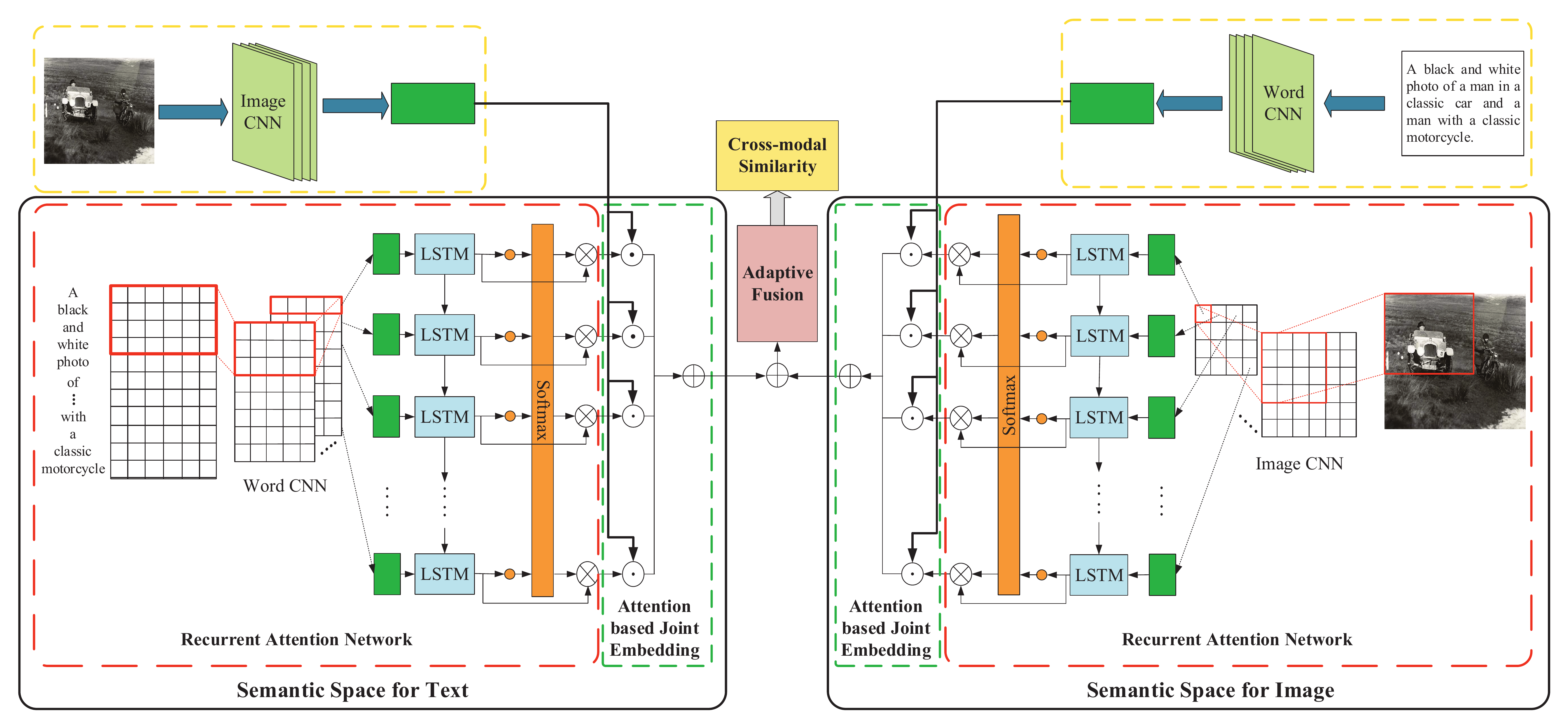}
	\end{minipage}%
	\setlength{\abovecaptionskip}{0.7cm}
	\caption{The overall framework of our MCSM approach, which adopts recurrent attention network with attention based joint embedding loss to construct independent semantic spaces for different modalities and perform cross-modal correlation learning, and the modality-specific cross-modal similarities are directly generated through end-to-end frameworks without explicit common representation.}\label{fig:network}
\end{figure*}

\subsection{Attention Mechanism}

Attention mechanism, as one of the recent advances in neural network, has been successfully applied to various multimodal tasks, which allows models to focus on the necessary fine-grained parts of visual or textual inputs. We mainly introduce two kinds of attention mechanism as follows, namely visual attention and textual attention.

\subsubsection{Visual attention}

Recently, many methods have adopted visual attention model to promote several image processing tasks, which can pay more attention to fine-grained parts in an image. Mnih et al. \cite{DBLP:conf/nips/MnihHGK14} propose an attention based task-driven visual processing framework for image classification, which adopts Recurrent Neural Network (RNN) to adaptively select a sequence of regions. Gregor et al. \cite{DBLP:conf/icml/GregorDGRW15} propose a spatial attention mechanism, which designs a sequential variational auto-encoding framework to perform image generation. Yang et al. \cite{DBLP:conf/cvpr/YangHGDS16} propose Stacked Attention Networks (SANs) for image question answering, which can locate relevant image regions to the question with stacked attention model.

\subsubsection{Textual attention}

Some related works in Natural Language Processing (NLP) have adopted textual attention model to find semantic alignments with an encoder-decoder network. Rockt{\"{a}}schel et al. \cite{DBLP:journals/corr/RocktaschelGHKB15} propose a word-by-word neural attention mechanism to reason over entailments of paired words or phrases. Hermann et al. \cite{DBLP:conf/nips/HermannKGEKSB15} develop a class of attention based deep neural networks, which learn to read and answer complex questions. Rush et al. \cite{DBLP:conf/emnlp/RushCW15} propose a fully data-driven approach, which adopts a local attention based model to generate summarization according to the input sentence.

Inspired by the great progress of attention mechanism, we attempt to fully exploit the intrinsic fine-grained information within each modality by attention mechanism, which can preserve the modality-specific characteristics when learning the cross-modal correlation.

\section{Our Proposed Approach}

As shown in Figure \ref{fig:network}, our proposed MCSM approach adopts modality-specific measurement to construct independent semantic spaces by an end-to-end framework for image and text respectively.
In each semantic space, first, \textit{recurrent attention network} is adopted to fully exploit the fine-grained modality-specific characteristics. Second, an \textit{attention based joint embedding} is employed to capture the imbalanced and complementary relationships between different modalities and perform cross-modal correlation learning. Finally, \textit{adaptive fusion} is proposed to explore the complementarity between different semantic spaces for cross-modal retrieval.
%where the fine-grained modality-specific characteristics are fully modeled by recurrent attention network, and an attention based joint embedding loss is employed to capture the imbalance relationships between different modalities and perform cross-modal correlation learning. Finally, adaptive fusion is adopted to explore the complementarity between different semantic space.

\subsection{Notation}

In the beginning, the formal definition of cross-modal retrieval is presented as follows. The two modalities involved in cross-modal retrieval are denoted as $I$ for image and $T$ for text. The multimodal dataset consists of two parts, namely training set and testing set. The training set is denoted as $D_{tr}=\{I_{tr},T_{tr}\}$. Image training data $I_{tr}=\{i_p\}_{p=1}^{n_{tr}}$ consists of totally $n_{tr}$ instances, and $i_p$ is the $p$-th image instance. Similarly, the text training data is denoted as $T_{tr}=\{t_p\}_{p=1}^{n_{tr}}$, which has the same number of instance with image for training. Besides, the training data also has its corresponding semantic category labels, which are denoted as $\{c_p^I\}_{p=1}^{n_{tr}}$ and $\{c_p^T\}_{p=1}^{n_{tr}}$. 
Then, the testing set, which is denoted as $D_{te}=\{I_{te},T_{te}\}$, includes $I_{te}=\{i_q\}_{q=1}^{n_{te}}$ and $T_{te}=\{t_q\}_{q=1}^{n_{te}}$ containing $n_{te}$ testing instances. 
Finally, given a query of any modality, the goal of cross-modal retrieval is to calculate the cross-modal similarity $sim(i_a,t_b)$, and retrieve the relevant instances of another modality in the testing data by the ranking of calculated similarity. 
In the following subsections, we first introduce the proposed image and text semantic space measurement methods respectively, then we demonstrate the proposed adaptive fusion approach on different semantic spaces.

\subsection{Image Semantic Space Measurement}

\subsubsection{Recurrent attention network for image}

To exploit intrinsic modality-specific characteristics within image data, we design a recurrent network with attention mechanism, which is adopted on the top of CNN hidden layer, to fully model the fine-grained local information and spatial context information jointly. 

%To exploit intrinsic modality-specific characteristics within image data, we design a Convolutational Recurrent Network (CNN-RNN) architecture, which adopts recurrent network on the top of CNN hidden layer with attention mechanism, to fully model the fine-grained local information and spatial context information jointly. 

First, each input image $i_p$ is resized to $256\times256$ and fed into CNNs for exploiting the fine-grained local information. Specifically, the network structure is configured the same as the layers before the last pooling layer (pool5) in 19-layer VGGNet \cite{DBLP:journals/corr/SimonyanZ14a}. The last pooling layer consists of 512 filters, and we can obtain separate feature vectors for different regions from the response of each filter over a $7\times7$ mapping of the image. 
Thus, each input image $i_p$ can be represented as $\{v^i_1,...,v^i_n\}$, where $n$ denotes the total number of image regions and $v^i_n$ is the $n$-th region represented as a 512 dimensional feature vector.

\begin{figure}[!t]
	\centering
	\includegraphics[width=0.4\textwidth]{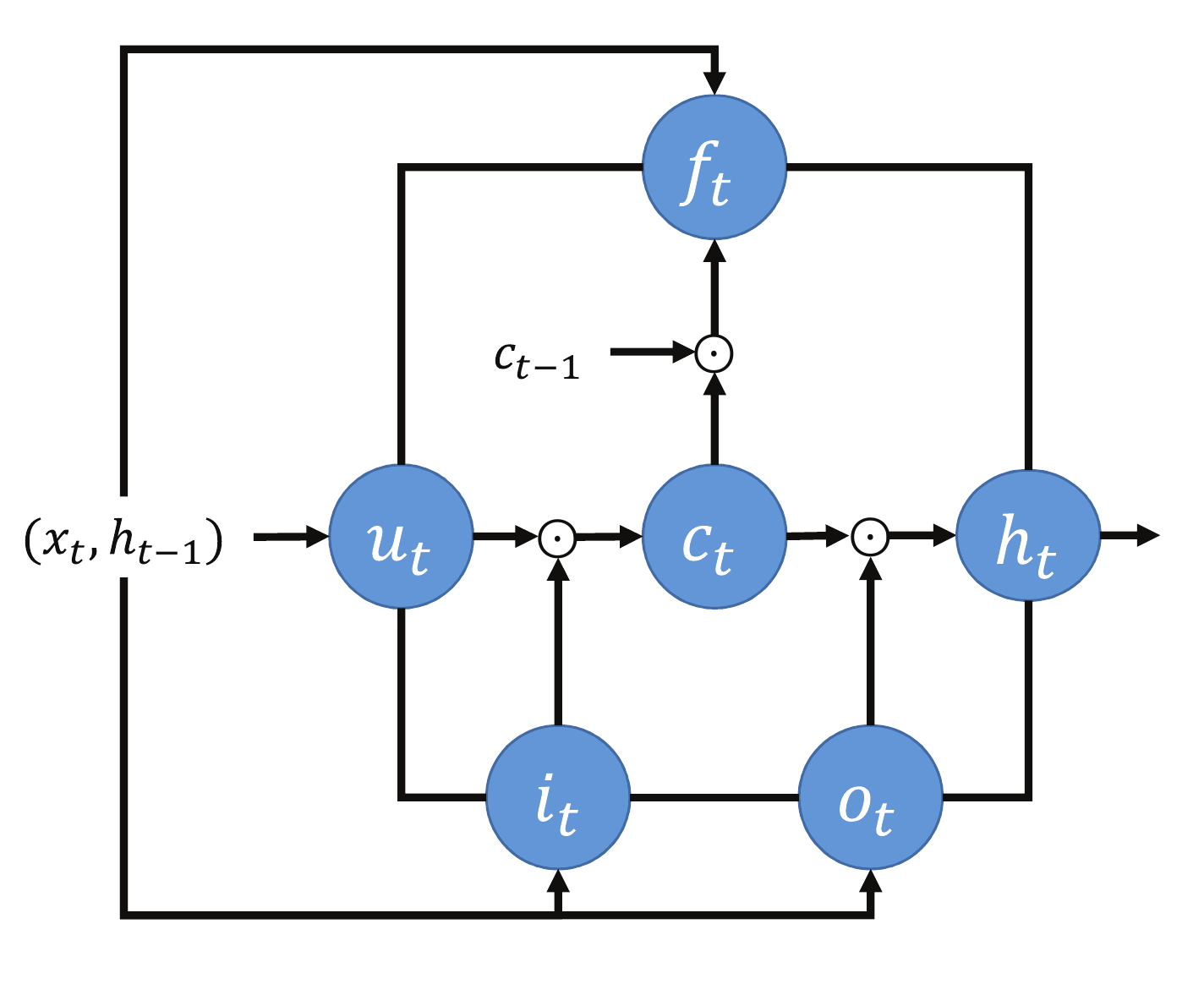}
	\setlength{\abovecaptionskip}{0.2cm}
	\caption{The architecture of an LSTM unit, which can learn long-term dependencies and retain information of previous time-steps through the memory cell and the update gates.
	}
	\label{fig_lstm}
\end{figure}

Then, we employ RNN to model the spatial context information among image regions. We compose these regions as a sequence, which are regarded as the results of eye movement when we glance at the image \cite{DBLP:journals/corr/LinPWZ17}. Specifically, Long Short Term Memory (LSTM) network \cite{DBLP:journals/neco/HochreiterS97} is adopted, which is a special kind of RNN, with strong ability to learn long-term dependencies through the memory cell and the update gates as well as preserve previous time-steps information at the same time. The architecture of LSTM unit is shown in Figure \ref{fig_lstm}. Formally, taking a sequence of image regions as input, the LSTM is updated recursively with the following equations.
\begin{align}
i_t&=\sigma(W_ix_t+U_ih_{t-1}+b_i), \label{l1}\\
f_t&=\sigma(W_fx_t+U_fh_{t-1}+b_f), \\
o_t&=\sigma(W_ox_t+U_oh_{t-1}+b_o), \\
u_t&=tanh(W_ux_t+U_uh_{t-1}+b_u), \\
c_t&=u_t\odot i_t+c_{t-1}\odot f_t, \\
h_t&=o_t\odot tanh(c_t), \label{l2}
\end{align}
where the activation vectors of the input, forget, memory cell and output are denoted as $i$, $f$, $c$ and $o$ respectively. And $x$ is the input region feature and the hidden unit output is denoted as $h$. While $W$, $U$ and $b$ are the weight matrices and bias term that need to be trained. $\odot$ denotes the element-wise multiplication. And $\sigma$ is the sigmoid nonlinearity to activate the gate, which is defined as follows.
\begin{align}
\sigma(x) = \frac{1}{1+e^{-x}}.
\end{align}

After that, we apply attention mechanism which aims to allow models focusing on the necessary fine-grained regions within image. We have obtained the output sequence $H_i=\{h^i_1,...,h^i_n\}$ from LSTM. The attention weights $a^i$ can be calculated by a feed-forward neural network with softmax function as follows.
\begin{align}
&M^i=tanh(W_a^iH_i), \\
&a^i=softmax(w_{ia}^TM_i),
\end{align}
where $W_a^i$ and $w_{ia}$ are the weight parameters for respective layers. And $a^i$ is the generated attention probabilities for image regions. Thus, the final image vector for the $n$-th region can be obtained as $a_n^ih^i_n$, which contains both fine-grained local information and spatial context information.

\subsubsection{Visual Attention based joint embedding}

Cross-modal correlation learning is performed to project the text data into the image semantic space, which utilizes the learned visual attention weights from the above recurrent attention network. 
We first need to generate the representation for text instance $t_p$. Each word is represented by a $k$-dimensional vector extracted by Word2Vec \cite{DBLP:conf/nips/MikolovSCCD13} model, which is trained on billions of words in Google News and publicly available\footnote{https://code.google.com/p/word2vec/}. Thus, a sentence with $n$ words can be represented as an $n\times k$ matrix. And a Word CNN is adopted on the input matrix, which has the same configuration with \cite{DBLP:conf/emnlp/Kim14}. While the text representation for each sentence is generated from the last fully connected layer, denoted as $q_p^t$.

Then, we aim to project the text data from its feature space into the constructed semantic space for image. To fully explore the imbalanced and complementary relationships between different modalities, we design the cross-modal similarity $sim_i$ for image semantic space between image $i_p$ and text $t_p$ as follows.
\begin{align}
sim_i(i_p,t_p)=\sum_{j=1}^{n}a_j^{i_p}h^{i_p}_j\cdot q_p^t, \label{sim_i}
\end{align}
where $h^{i_p}_j$ is the $j$-th region in the image $i_p$ and $a_j^{i_p}$ is the attention weight for the corresponding image region. 

Finally, we design an attention based joint embedding loss to perform cross-modal correlation learning, which jointly considers both matched and mismatched image/text pairs with the defined cross-modal similarity based on the learned attention weights. The objective function is defined as follows.
\begin{align}
L_i=\frac{1}{N}\sum_{n=1}^{N}l_{i1}(i_n^{+},t_n^{+},t_n^{-}) + l_{i2}(t_n^{+},i_n^{+},i_n^{-}),
\end{align}
and the two items in this formula are defined as:
\begin{align}
l_{i1}(i_n^{+},&t_n^{+},t_n^{-})= \notag \\&max(0,\alpha +sim_i(i_n^{+},t_n^{+})-sim_i(i_n^{+},t_n^{-})), \label{li1}\\
l_{i2}(t_n^{+},&i_n^{+},i_n^{-})= \notag \\&max(0,\alpha +sim_i(i_n^{+},t_n^{+})-sim_i(i_n^{-},t_n^{+})), \label{li2}
\end{align}
where $(i_n^{+},t_n^{+})$ denotes the matched image pair, while $(i_n^{+},t_n^{-})$ and $(i_n^{-},t_n^{+})$ are the mismatched pairs. The margin parameter is set to be $\alpha$. $N$ is the number of triplet tuples sampled from training set. To train the model, stochastic gradient descent (SGD) is adopted. 
%Note that network weights of convolutional layer for image are fixed to the pre-trained model, while the gradients are backpropagated through all other network parameters. 
So far, we have obtained the modality-specific cross-modal similarity $sim_i$ for image semantic space, which integrates both representation learning and similarity measurement to benefit each other, and also fully captures the intrinsic fine-grained clues in image and imbalanced information across different modalities for correlation learning.

\subsection{Text Semantic Space Measurement}

\subsubsection{Recurrent attention network for text}

For fully exploiting modality-specific characteristics within text data, we also design a recurrent network with attention mechanism, on the top of CNN hidden layer, which can model both the fine-grained local and context information in textual descriptions.

First, the input text instance $t_p$, which consists of $n$ words, is represented as an $n\times k$ matrix, where each word is also represented as a $k$-dimensional vector extracted by Word2Vec \cite{DBLP:conf/nips/MikolovSCCD13} model. Following \cite{DBLP:conf/emnlp/Kim14}, we design fast convolutional networks for text (Word CNN), which is built by several combinations of convolution layer, threshold activation function layer and pooling layer. The Word CNN is similar with the image CNN except the 2D convolution and spatial max-pooling of image CNN are replaced by temporal (1D) convolution and temporal max-pooling. Through the Word CNN network, CNN hidden activation of the last pooling layer is split to generate the features of text fragments.
%we can get the features of text fragments in order from the last pooling layer. 

Second, RNN is adopted on the top of CNN with the sequence of vectors to further model the context information along the input text sequence. Specifically, we also adopt LSTM network \cite{DBLP:journals/neco/HochreiterS97} to exploit such temporal dependency, which takes a sequence of text fragments as input. LSTM is updated following the equations (\ref{l1}) to (\ref{l2}), where $x$ denotes the input feature of text fragment. Thus, we can obtain the output sequence from LSTM denoted as $H_t=\{h^t_1,...,h^t_m\}$.

Then, the attention mechanism is applied to capture useful fine-grained fragments in text sequence. The attention weights are denoted as $a^t$, which are calculated by a feed-forward network with softmax function as follows. 
\begin{align}
&M^t=tanh(W_a^tH_t), \\
&a^t=softmax(w_{ta}^TM_t),
\end{align}
where the weight parameters for respective layers are denoted as $W_a^t$ and $w_{ia}$. $a^t$ is the generated attention probabilities for text fragments. Thus, the final text vector for the $m$-th fragment is calculated as $a_m^th^t_m$, which captures both fine-grained local and context information in textual description. 

\subsubsection{Textual Attention based joint embedding}

To perform cross-modal correlation learning, image data is projected into the text semantic space to utilize the learned textual attention weights from the above recurrent attention network for text. 
We still need to generate the image representation for each instance $i_p$, which is also extracted from the last fully connected layer (fc7) in 19-layer VGGNet \cite{DBLP:journals/corr/SimonyanZ14a} with 4,096 dimensional number, and denoted as $q_p^i$.

Then, the image data is projected into the constructed semantic space for text from their own feature space. Thus, we design the cross-modal similarity $sim_t$ for text semantic space between image $i_p$ and text $t_p$ as follows, which aims to explore the imbalanced and complementary relationships between different modalities.
\begin{align}
sim_t(i_p,t_p)=\sum_{j=1}^{m}a_j^{t_p}h^{t_p}_j\cdot q_p^i,\label{sim_t}
\end{align}
where the $j$-th fragment of text $t_p$ is denoted as $h^{t_p}_j$, $a_j^{t_p}$ is the attention weight for the corresponding text fragment. 

Finally, an attention based joint embedding loss is designed similarly for performing cross-modal correlation learning in the text semantic space, which takes the consideration that the difference between the similarity of matched image/text pair and that of mismatched pair should be as large as possible. Thus, the objective function is defined as follows.
\begin{align}
L_t=\frac{1}{M}\sum_{n=1}^{M}l_{t1}(t_n^{+},i_n^{+},i_n^{-})+l_{t2}(i_n^{+},t_n^{+},t_n^{-}),
\end{align}
where $l_{t1}(t_n^{+},i_n^{+},i_n^{-})$ and $l_{t2}(i_n^{+},t_n^{+},t_n^{-})$ are defined similarly as equations (\ref{li1}) and (\ref{li2}) with the cross-modal similarity $sim_t$ as follows, 
\begin{align}
l_{t1}(t_n^{+},&i_n^{+},i_n^{-})= \notag \\&max(0,\beta +sim_t(i_n^{+},t_n^{+})-sim_t(i_n^{-},t_n^{+})), \label{lt1} \\
l_{t2}(i_n^{+},&t_n^{+},t_n^{-})= \notag \\&max(0,\beta +sim_t(i_n^{+},t_n^{+})-sim_t(i_n^{+},t_n^{-})). \label{lt2} 
\end{align}
Also the number of triplet tuples sampled from training set is denoted as $M$. $\beta$ is the margin parameter. 
%The network weights of convolutional layer in image CNN is still fixed during training stage. And SGD is adopted to backpropagate the gradients through all other network parameters. 
Therefore, the modality-specific cross-modal similarity $sim_t$ for text semantic space can be obtained, with the fully modeling of fine-grained clues in text as well as imbalanced information across different modalities for correlation learning. 

\subsection{Adaptive Fusion on Different Semantic Spaces}

So far, we have obtained two kinds of modality-specific cross-modal similarity, namely $sim_i$ and $sim_t$ from the semantic spaces for image and text. Inspired by \cite{DBLP:journals/corr/KongDDZT16}, we further attempt to explore the complementarity between different semantic spaces by adaptive fusion.

First, the cross-modal similarity scores obtained from different semantic spaces are min-max normalized to $[0,1]$ respectively, which aims to overcome the influence of image/text pairs with too large similarity scores, and are defined as follows. 
\begin{align}
r_i(i_p,t_p)=\frac{sim_i(i_p,t_p)-min(sim_i(i,t))}{max(sim_i(i,t))-min(sim_i(i,t))}, \\
r_t(i_p,t_p)=\frac{sim_t(i_p,t_p)-min(sim_t(i,t))}{max(sim_t(i,t))-min(sim_t(i,t))}.
\end{align}
Then, the normalized scores obtained from one semantic space are used as the adaptive weights of the corresponding image/text pair in another semantic space during fusion stage, with the motivation that larger similarity in one semantic space leads to higher importance of the corresponding pair in another semantic space.  
Finally, the two kinds of modality-specific cross-modal similarity are fused with the following equation.
\begin{align}
Sim(&i_p,t_p)=\\&r_t(i_p,t_p)\cdot sim_i(i_p,t_p)+r_i(i_p,t_p)\cdot sim_t(i_p,t_p).\notag 
\end{align}
Thus, we can obtain the final cross-modal similarity score $Sim(i_p,t_p)$ between image $i_p$ and text $t_p$, which can fully explore the complementarity between different semantic spaces to boost the performance of cross-modal retrieval.

\section{Experiments}

In this section, we conduct experiments on 3 cross-modal datasets, including widely-used Wikipedia and Pascal Sentence datasets as well as our constructed large-scale XMediaNet dataset, taking 9 state-of-the-art methods for comparison to verify the effectiveness of our proposed approach. Besides, comprehensive experimental analyses are presented including convergence and parameter analysis, as well as baseline experiments to verify the contribution of each component in our approach.

\begin{figure}[!t]
	\centering
	\includegraphics[width=0.485\textwidth]{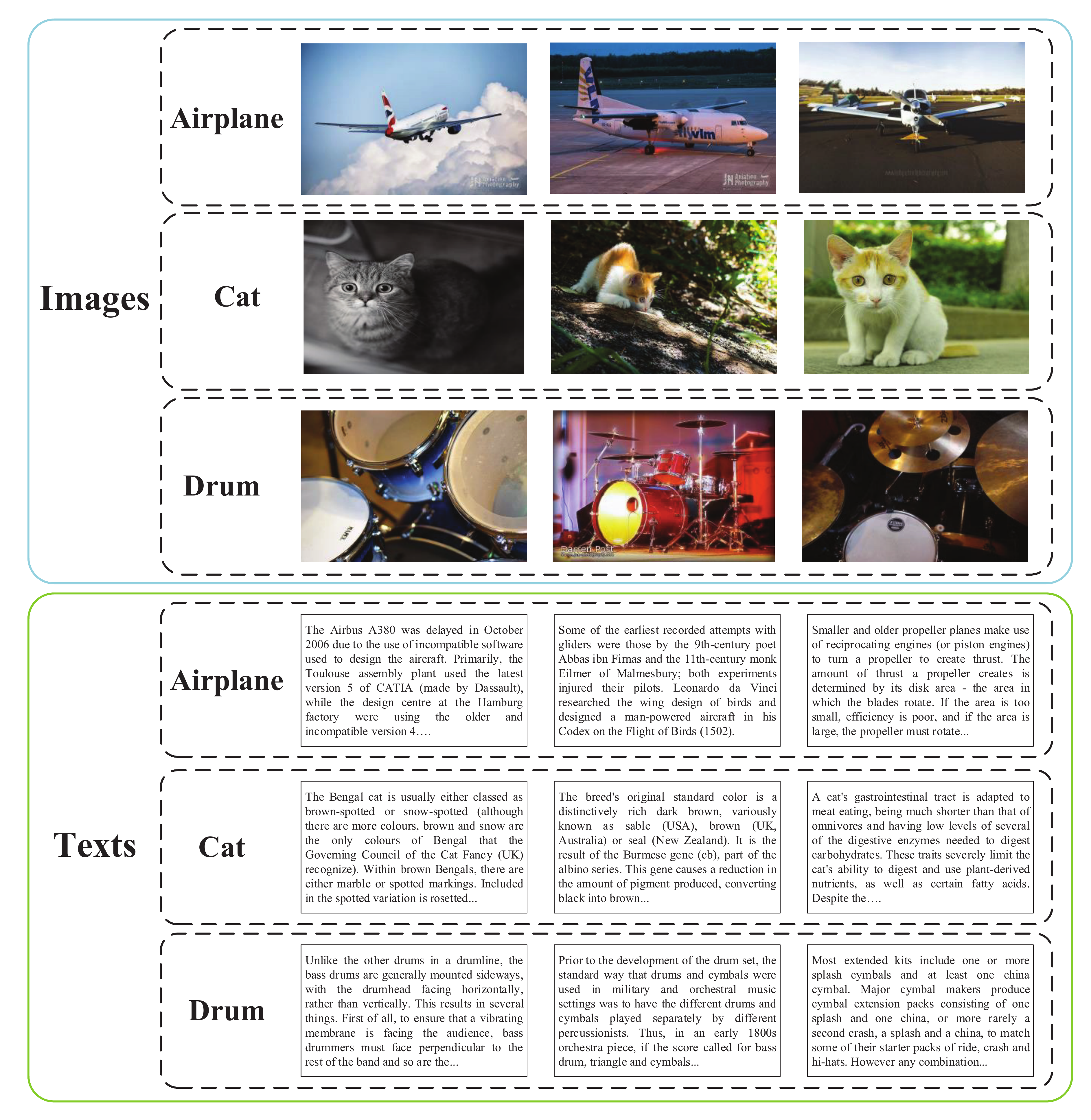}
	\setlength{\abovecaptionskip}{0.2cm}
	\caption{Examples of 3 categories from XMediaNet dataset, which are airplane, cat and drum.}
	\label{fig_wiki_ex}
\end{figure}

\subsection{Datasets}

Here we briefly introduce 3 cross-modal datasets adopted in the experiments, including Wikipedia, Pascal Sentence and our constructed large-scale XMediaNet datasets. Each dataset is divided into 3 subsets, namely training set, testing set and validation set. 
%For fair and objective comparison purpose, we exactly follow the dataset partition strategy of \cite{feng12014cross,DBLP:conf/ijcai/PengHQ16} in the experiment to divide the dataset into 3 subsets: Training set, testing set and validation set. 
\begin{itemize}
	\item { \textbf{Wikipedia dataset}} \cite{RasiwasiaMM10SemanticCCA}, as the most widely-used dataset for cross-modal retrieval, is selected from ``featured articles'' in Wikipedia\footnote{https://en.wikipedia.org/wiki/Wikipedia:Featured\_articles} with 10 most populated categories, including history, biology and so on. This dataset totally consists of 2,866 image/text pairs. For fair and objective comparison purpose, we exactly follow the dataset partition strategy of \cite{DBLP:conf/ijcai/PengHQ16,feng12014cross} to divide the dataset into 3 subsets: 2,173 pairs in training set, 231 pairs in validation set and 462 pairs in testing set.
	
	\item {  \textbf{Pascal Sentence dataset}} \cite{rashtchian2010collecting} contains 1,000 images, which is generated from 2008 PASCAL development kit. Each image is annotated via Amazon Mechanical Turk by crowdsourcing to generate 5 independent sentences from different annotators, which forms one document. This dataset is categorized into 20 categories, and also following \cite{DBLP:conf/ijcai/PengHQ16,feng12014cross}, 800 documents are selected as training set, while 100 documents for testing and 100 documents for validation.
	
	\item {  \textbf{XMediaNet dataset}} is our constructed large-scale cross-modal dataset, which consists of 5 modalities, namely text, image, video, audio and 3D model. We select 200 category nodes from WordNet\footnote{http://wordnet.princeton.edu/} to construct this dataset to ensure the semantic hierarchy structure. These categories can be divided into two main kinds: animals and artifacts. There are 47 species of animal such as elephant, owl, bee and frog as well as 153 types of artifact such as violin, airplane, shotgun, and camera. The total number of instances exceeds 100,000. It is noted that we also use image and text in the experiments, where text paragraphs are extracted from Wikipedia articles whose topics belong to the category, and images including the objects of the category are collected from Flickr\footnote{http://www.flickr.com}. This dataset totally consists of 40,000 image/pairs, and is also divided into 3 subsets, with 32,000 pairs in training set, 4,000 pairs in testing set and 4,000 pairs in validation set. Some examples are shown in Figure \ref{fig_wiki_ex}.
\end{itemize}

\subsection{Details of the Network}

We implement the proposed network by Torch\footnote{http://torch.ch/}, which is a widely-used scientific computing framework. And we introduce the details of the network including data preprocess strategy and network structure as following.

\subsubsection{Data preprocess}

For image, we use the original images resized to $256\times256$ as inputs. For text, we convert each word in a document into a 300-dimension vector by Word2Vec \cite{DBLP:conf/nips/MikolovSCCD13} and generate vector sequences as text inputs. 
The maximum input length is set as the max sequence length in the dataset, and we adopt zero-padding for other sequences to beneath this limit.
%We set the max sequence length as the input length and pad the sequences which beneath this limit with zero.

\subsubsection{Recurrent attention network}

The recurrent attention network of each semantic space mainly consists of three parts, namely convolutional network, Long Short-Term Memory (LSTM) network and attention network. For the semantic space for text, the convolutional network consists of three learnable temporal convolution layers, each of which is followed by a ReLU activation function layer and a temporal max-pooling layer. Taking Wikipedia dataset as an example, the first temporal convolution layer has 384 kernels whose widths are 15. The parameter combinations of the remaining convolution layers are (512, 9) and (256, 7). The first parameter of each combination means the number of convolution kernels while the second is the kernel width. The kernel step sizes of all convolution layers are 1. For the other two datasets, the number of kernels on each convolution layer is the same with Wikipedia dataset, but the length of text instances differs greatly between different datasets, thus the kernel widths change according to the lengths of the input sequences. For the semantic space for image, we use the pretrained convolution network in 19-layer VGGNet and treat the output of the last pooling layer as the input of LSTM. The LSTM network has two units in series, and the dimension of output keeps the same with input. The LSTM is followed by a fully connected layer which aims to project the output of LSTM into the target dimension, which is 4,096 dimensions in our case. The attention network is made up of a fully connected layer and a softmax layer. 

\subsection{Compared Methods}

Totally 9 state-of-the-art methods are compared in the experiments to verify the effectiveness of our proposed approach. There are 5 traditional cross-modal retrieval methods, namely CCA \cite{HotelingBiometrika36RelationBetweenTwoVariates}, CFA \cite{LiMM03CFA}, KCCA \cite{DBLP:journals/neco/HardoonSS04}, JRL \cite{ZhaiTCSVT2014JRL} and LGCFL \cite{DBLP:journals/tmm/KangXLXP15}. While the other 4 methods, Corr-AE \cite{feng12014cross}, DCCA \cite{DBLP:conf/cvpr/YanM15}, CMDN \cite{DBLP:conf/ijcai/PengHQ16} and Deep-SM \cite{DBLP:journals/tcyb/WeiZLWLZY17} are DNN-based methods.
The brief introductions of these 9 compared methods are presented as follows:
\begin{itemize}
	\item {\bf CCA} \cite{HotelingBiometrika36RelationBetweenTwoVariates} learns project matrices to maximize the correlation between the projected features of different modalities in one common space.
	
	\item {\bf CFA} \cite{LiMM03CFA} minimizes the Frobenius norm between the data of different modalities after projecting them into one common space.
	
	\item {\bf KCCA} \cite{DBLP:journals/neco/HardoonSS04} uses kernel function to project the features into a higher-dimensional space, and then learns a common space by CCA. In the experiments, we adopt Gaussian kernel as the kernel function.
	
	\item {\bf JRL} \cite{ZhaiTCSVT2014JRL} learns a common space by using semantic information, with semi-supervised regularization and sparse regularization.
	
	\item {\bf LGCFL} \cite{DBLP:journals/tmm/KangXLXP15} jointly learns basis matrices of different modalities, by using a local group based priori in the formulation to fully take advantage of popular block based features.

	\item {\bf Corr-AE} \cite{feng12014cross} consists of two autoencoder networks coupled at the code layer to simultaneously model the reconstruction error and correlation loss. It should be noted that Corr-AE has two extensions, namely Corr-Cross-AE and Cross-Full-AE, and in the experiments we compare with the best results of the three models.
	
	\item {\bf DCCA} \cite{DBLP:conf/cvpr/YanM15} is a nonlinear extension of CCA. The correlation is maximized between the output layers of two separate subnetworks.
	
	\item {\bf CMDN} \cite{DBLP:conf/ijcai/PengHQ16} adopts multiple deep networks to generate separate representations and learns the common representation with a stacked network.
	
	\item {\bf Deep-SM} \cite{DBLP:journals/tcyb/WeiZLWLZY17} adopts convolutional neural network to perform deep semantic matching, which fully exploits the strong power of CNN image feature.

\end{itemize}

\subsection{Evaluation Metric}

We perform cross-modal retrieval on 3 datasets mentioned above with two kinds of retrieval tasks, which are defined as follows.
\begin{itemize}
	\item {\textbf{Image query text}} (image$\to$text). Retrieving relevant text instances in the testing set ranked by calculated cross-modal similarity, using a query of image.
	\item {\textbf{Text query image}} (text$\to$image). Retrieving relevant image instances in the testing set ranked by calculated cross-modal similarity, using a query of text.
\end{itemize}

It should be noted that our proposed MCSM approach integrates both common representation learning and distance metric, which takes original image and text as inputs to directly generate the cross-modal similarity score. While other compared methods only learn the common representation taking hand-crafted features as input. Thus, for fair comparison, all compared methods also adopt the same CNN features used in our proposed approach as input. Specifically, the CNN feature of image is extracted from fc7 layer in 19-layer VGGNet \cite{DBLP:journals/corr/SimonyanZ14a}, while CNN feature of text is extracted by Word CNN with the same configuration of \cite{DBLP:conf/emnlp/Kim14}.
We directly adopt the source codes provided by their authors, to fairly evaluate the compared methods by the following steps in the experiments. 

\textit{1)} Perform common representation learning using the data in training set to obtain the learned projections or deep models.

\textit{2)} Use the learned projections or deep models to convert the data in testing set into the common representation.

\textit{3)} Calculate the cross-modal similarity between image and text by cosine distance, and then perform cross-modal retrieval.

Mean Average Precision (MAP) score is adopted as the evaluation metric on Wikipedia, Pascal Sentence and our constructed XMediaNet datasets, which is the mean value of Average Precision (AP) of each query. And AP is defined as follows.
\begin{align}
AP= \frac{1}{R}\sum_{k=1}^{n}\frac{R_{k}}{k}\times rel_{k},
\end{align}
where the testing set contains $n$ instances, which has $R$ relevant instances. $R_k$ is the number of relevant instances in the top $k$ returned results. $rel_k$ is set to be 1 when the $k$-th returned result is relevant, otherwise, $rel_k$ is set to be 0. 
MAP score considers the ranking of returned retrieval results as well as precision simultaneously, which is extensively adopted in cross-modal retrieval tasks, such as \cite{RasiwasiaMM10SemanticCCA,DBLP:conf/ijcai/PengHQ16}. It should be noted that the MAP score is calculated on \emph{all returned results}, not only \textit{top 50} as adopted in \cite{feng12014cross}. %Furthermore, precision-recall and precision-scope curves on large-scale XMediaNet dataset are adopted for more comprehensive comparison.

\begin{table}[htb]
	\caption{The MAP scores of Cross-modal Retrieval for our MCSM approach and 9 compared methods on \textbf{Wikipedia} dataset.}
	\begin{center}
		\scalebox{1.0}{
			\begin{tabular}{|c|c|c|c|} 
				\hline
				\multirow{2}{*}{Method} & \multicolumn{3}{c|}{MAP scores} \\
				\cline{2-4}
				& Image$\rightarrow$Text & Text$\rightarrow$Image & Average \\
				\hline
				\textbf{Our MCSM Approach} & \textbf{0.516} & \textbf{0.458} & \textbf{0.487} \\
				CMDN \cite{DBLP:conf/ijcai/PengHQ16} & 0.487 & 0.427 & 0.457  \\
				Deep-SM \cite{DBLP:journals/tcyb/WeiZLWLZY17} & 0.478 & 0.422 & 0.450  \\
				LGCFL \cite{DBLP:journals/tmm/KangXLXP15} & 0.466 & 0.431 & 0.449\\
				JRL \cite{ZhaiTCSVT2014JRL} & 0.479 & 0.428& 0.454 \\
				DCCA \cite{DBLP:conf/cvpr/YanM15} & 0.445 & 0.399& 0.422 \\
				Corr-AE \cite{feng12014cross} & 0.442 & 0.429& 0.436 \\
				KCCA \cite{DBLP:journals/neco/HardoonSS04} & 0.438 & 0.389& 0.414 \\
				CFA \cite{LiMM03CFA} & 0.319 & 0.316& 0.318 \\
				CCA \cite{HotelingBiometrika36RelationBetweenTwoVariates} & 0.298 & 0.273& 0.286  \\
				\hline
				
			\end{tabular} 
		}
	\end{center}
	%\vspace{-0.5cm}
	\label{table:wiki}
\end{table}

\begin{table}[htb]
	\caption{The MAP scores of Cross-modal Retrieval for our MCSM approach and 9 compared methods on \textbf{Pascal Sentence} dataset.}
	\begin{center}
		\scalebox{1.0}{
			\begin{tabular}{|c|c|c|c|} 
				\hline
				\multirow{2}{*}{Method} & \multicolumn{3}{c|}{MAP scores} \\
				\cline{2-4}
				& Image$\rightarrow$Text & Text$\rightarrow$Image & Average \\
				\hline
				\textbf{Our MCSM Approach} & \textbf{0.598} & \textbf{0.598} & \textbf{0.598} \\
				CMDN \cite{DBLP:conf/ijcai/PengHQ16} & 0.544 & 0.526 & 0.535  \\
				Deep-SM \cite{DBLP:journals/tcyb/WeiZLWLZY17} & 0.560 & 0.539 & 0.550  \\
				LGCFL \cite{DBLP:journals/tmm/KangXLXP15} & 0.539 & 0.503 & 0.521\\
				JRL \cite{ZhaiTCSVT2014JRL} & 0.563 & 0.505& 0.534 \\
				DCCA \cite{DBLP:conf/cvpr/YanM15} & 0.568 & 0.509& 0.539 \\
				Corr-AE \cite{feng12014cross} & 0.532 & 0.521& 0.527 \\
				KCCA \cite{DBLP:journals/neco/HardoonSS04} & 0.488 & 0.446& 0.467 \\
				CFA \cite{LiMM03CFA} & 0.476 & 0.470& 0.473 \\
				CCA \cite{HotelingBiometrika36RelationBetweenTwoVariates} & 0.203 & 0.208& 0.206  \\
				\hline
				
			\end{tabular} 
		}
	\end{center}
	%\vspace{-0.5cm}
	\label{table:pas}
\end{table}

\begin{table}[htb]
	\caption{The MAP scores of Cross-modal Retrieval for our MCSM approach and 9 compared methods on \textbf{XMediaNet} dataset.}
	\begin{center}
		\scalebox{1.0}{
			\begin{tabular}{|c|c|c|c|} 
				\hline
				\multirow{2}{*}{Method} & \multicolumn{3}{c|}{MAP scores} \\
				\cline{2-4}
				& Image$\rightarrow$Text & Text$\rightarrow$Image & Average \\
				\hline
				\textbf{Our MCSM Approach} & \textbf{0.540} & \textbf{0.550} & \textbf{0.545} \\
				CMDN \cite{DBLP:conf/ijcai/PengHQ16} & 0.485 & 0.516 & 0.501  \\
				Deep-SM \cite{DBLP:journals/tcyb/WeiZLWLZY17} & 0.399 & 0.342 & 0.371  \\
				LGCFL \cite{DBLP:journals/tmm/KangXLXP15} & 0.441 & 0.509 & 0.475\\
				JRL \cite{ZhaiTCSVT2014JRL} & 0.488 & 0.405& 0.447 \\
				DCCA \cite{DBLP:conf/cvpr/YanM15} & 0.425 & 0.433& 0.429 \\
				Corr-AE \cite{feng12014cross} & 0.469 & 0.507& 0.488 \\
				KCCA \cite{DBLP:journals/neco/HardoonSS04} & 0.252 & 0.270& 0.261 \\
				CFA \cite{LiMM03CFA} & 0.252 & 0.400& 0.326 \\
				CCA \cite{HotelingBiometrika36RelationBetweenTwoVariates} & 0.212 & 0.217& 0.215  \\
				\hline
				
			\end{tabular} 
		}
	\end{center}
	%\vspace{-0.5cm}
	\label{table:xmn}
\end{table}

\begin{figure*}[!t]
	\centering
	\includegraphics[width=0.86\textwidth]{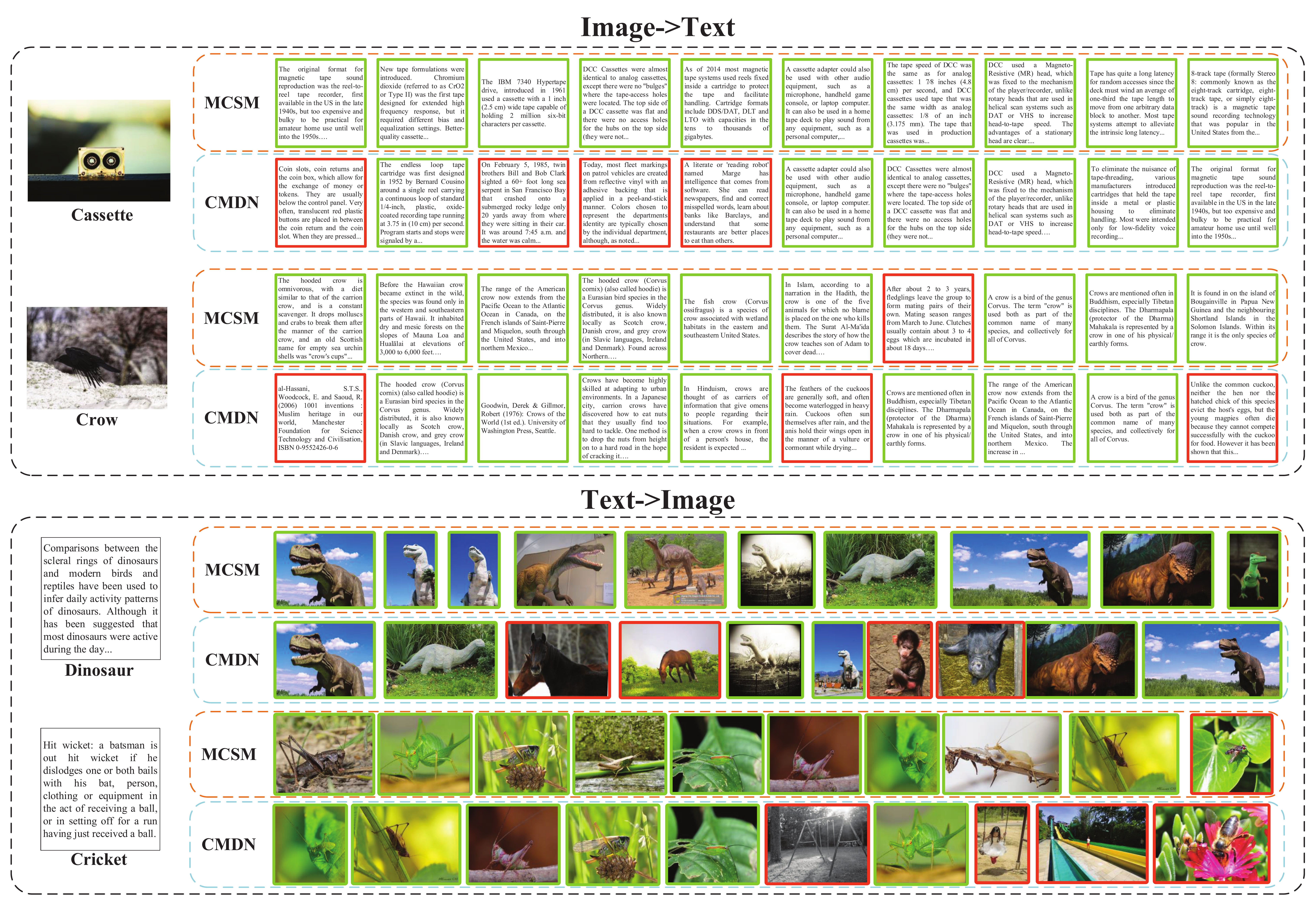}
	\setlength{\abovecaptionskip}{0.4cm}
	\caption{Examples of the cross-modal retrieval results on XMediaNet dataset by our proposed MCSM approach as well as compared method CMDN \cite{DBLP:conf/ijcai/PengHQ16}. In these examples, the correct retrieval results are with green borders, and the wrong results are with red borders.
	}
	\setlength{\abovecaptionskip}{0.5cm}
	\label{fig_xmn_res}
\end{figure*}

\begin{figure}[!t]
	\centering
	\includegraphics[width=0.43\textwidth]{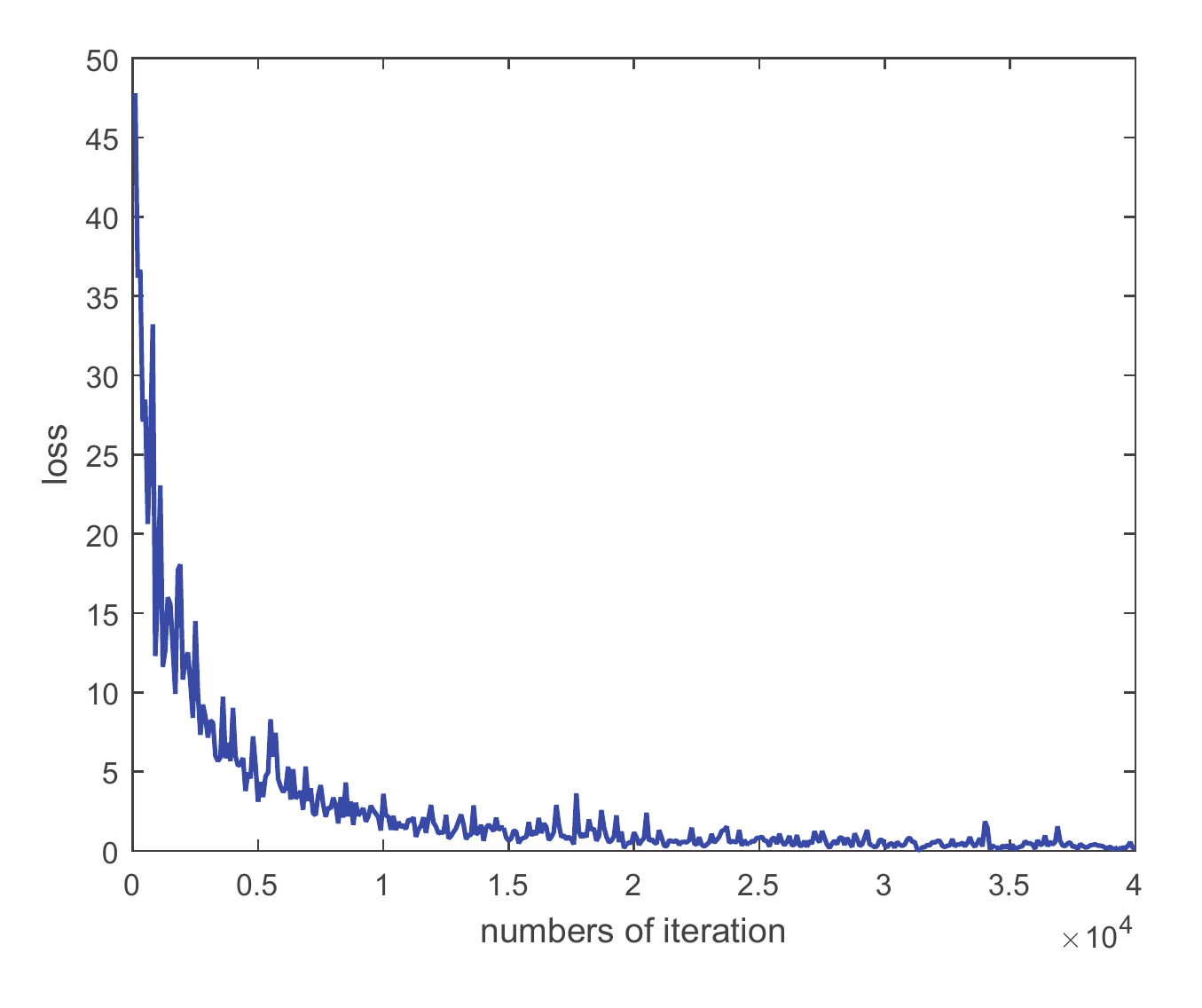}
	\setlength{\abovecaptionskip}{0.0cm}
	\caption{Convergence experiments of our proposed MCSM approach conducted on the large-scale XMediaNet dataset, which show the curves of downtrend on the loss value.
	}
	\label{fig_loss}
\end{figure}

\begin{figure*}[!t]
	
	\begin{center}
		
		\subfigure[Wikipedia dataset]
		{\includegraphics[width=0.31\textwidth]{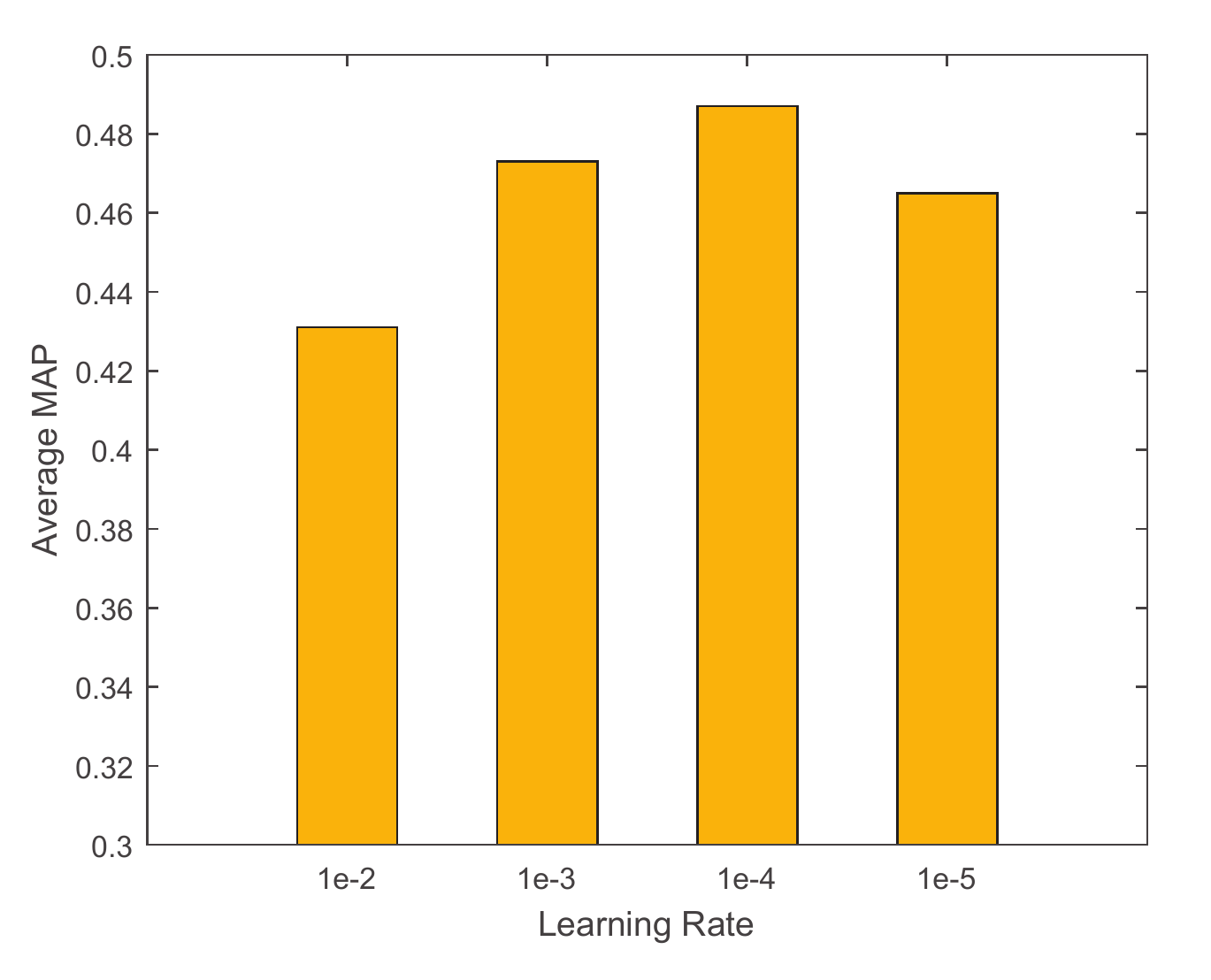}}
		\subfigure[Pascal Sentence dataset]
		{\includegraphics[width=0.31\textwidth]{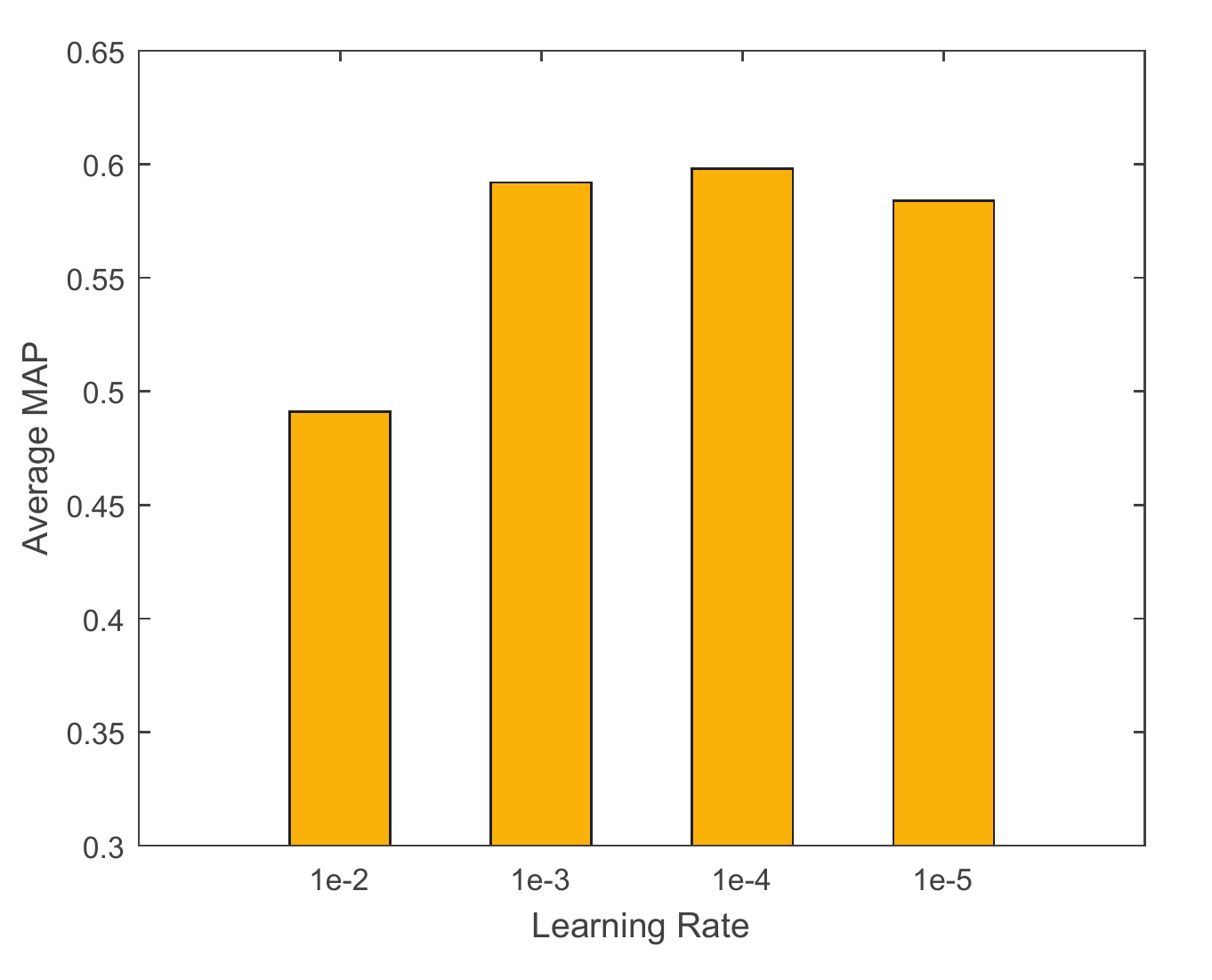}}
		\subfigure[XMediaNet dataset]
		{\includegraphics[width=0.31\textwidth]{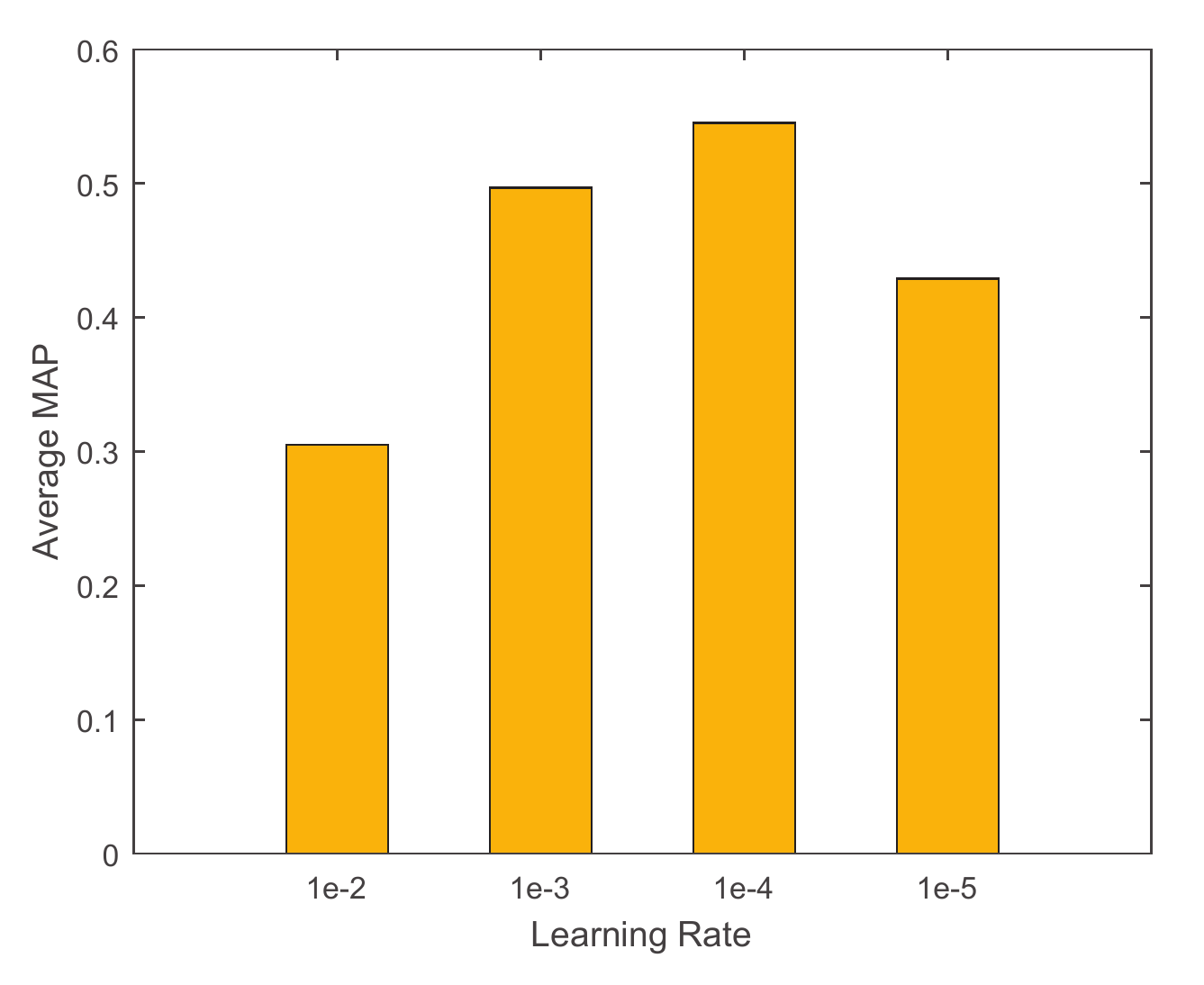}}
		\setlength{\abovecaptionskip}{0.3cm}
		\caption{Experiments on the influence of the learning rate, on Wikipedia, Pascal Sentence and XMediaNet datasets. It should be noted that we report the average MAP score of Image$\rightarrow$Text and Text$\rightarrow$Image tasks.
		}
		\label{fig_lr}
	\end{center}
	
\end{figure*}

\begin{figure*}[!t]
	
	\begin{center}
		
		\subfigure[Wikipedia dataset]
		{\includegraphics[width=0.31\textwidth]{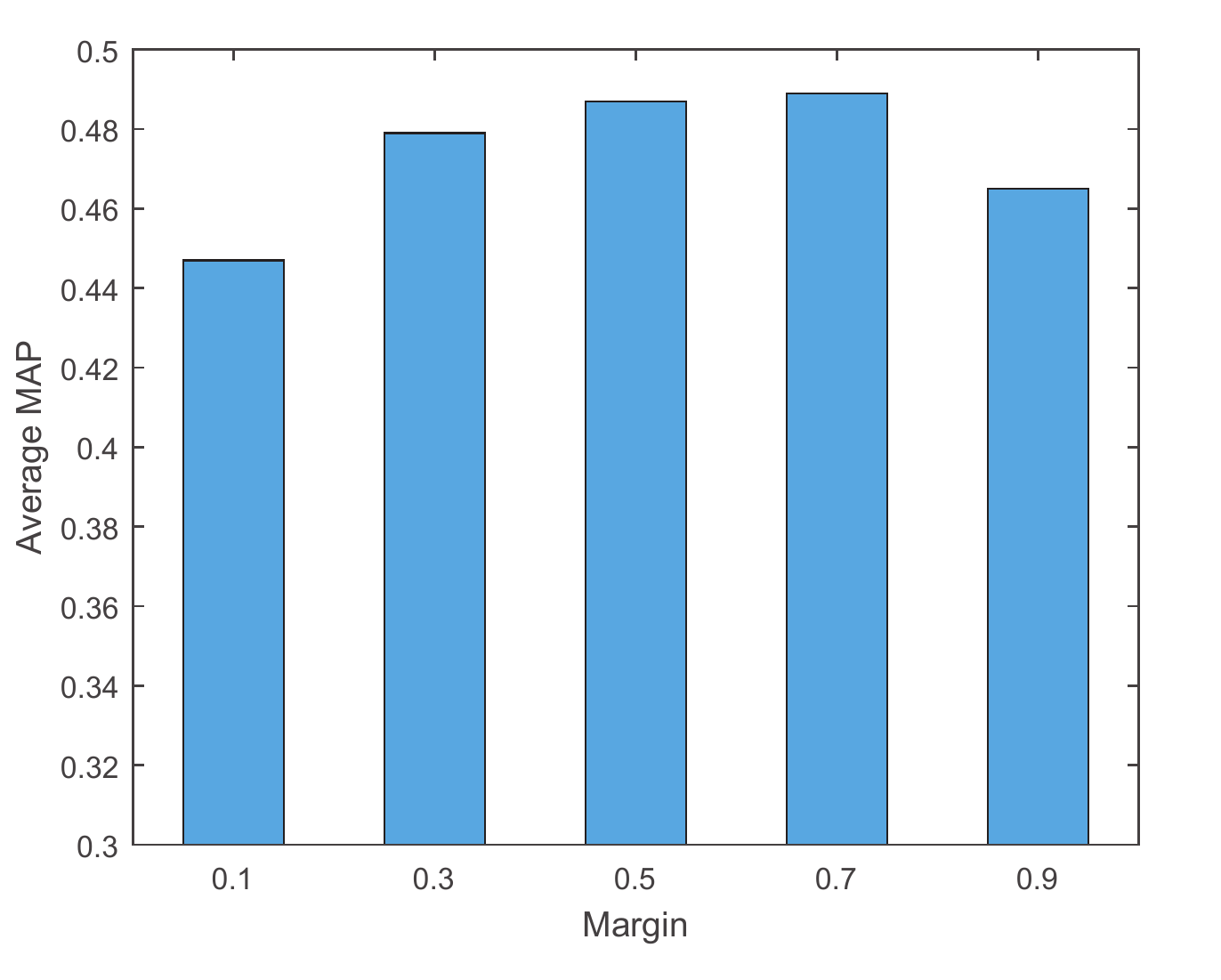}}
		\subfigure[Pascal Sentence dataset]
		{\includegraphics[width=0.31\textwidth]{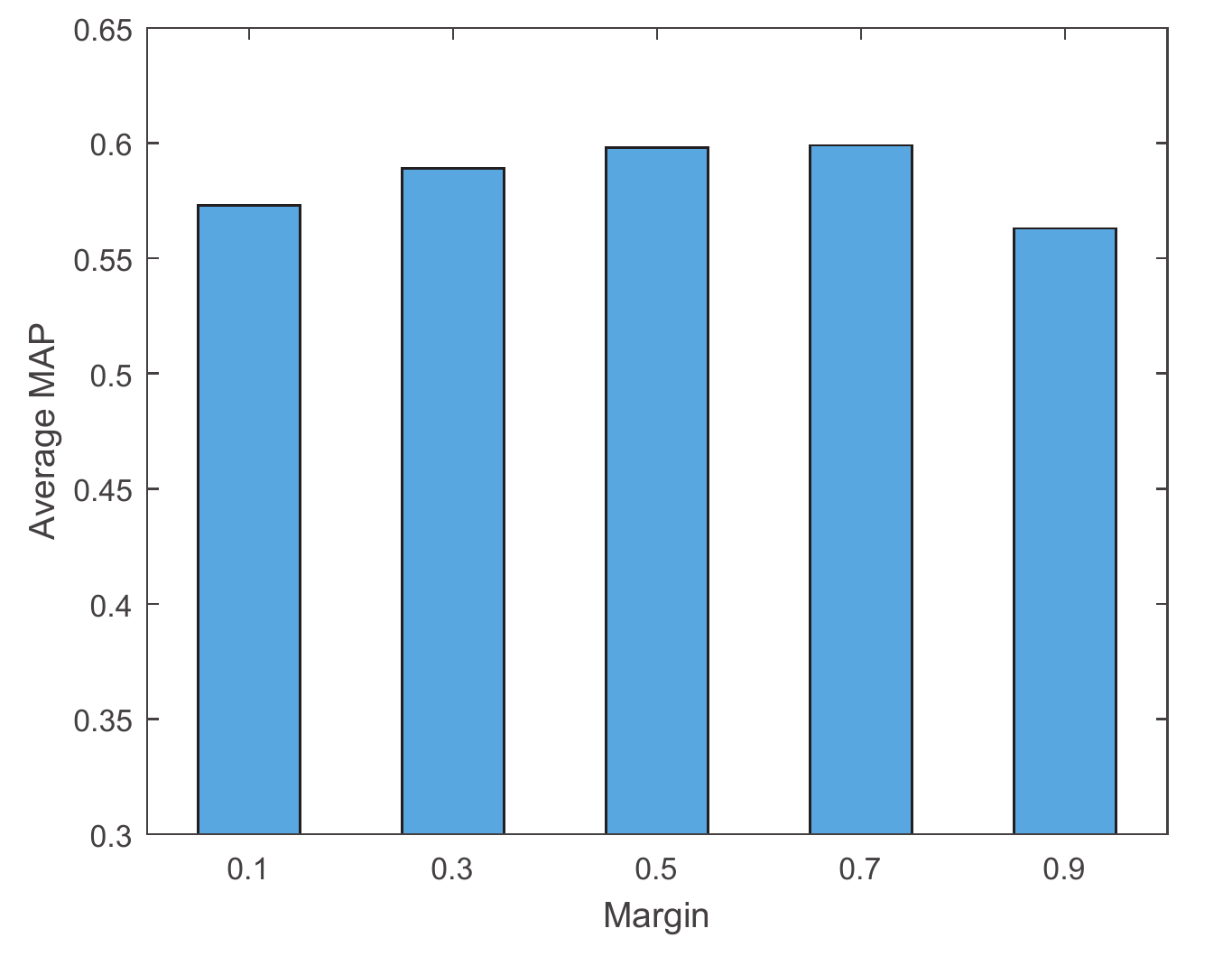}}
		\subfigure[XMediaNet dataset]
		{\includegraphics[width=0.31\textwidth]{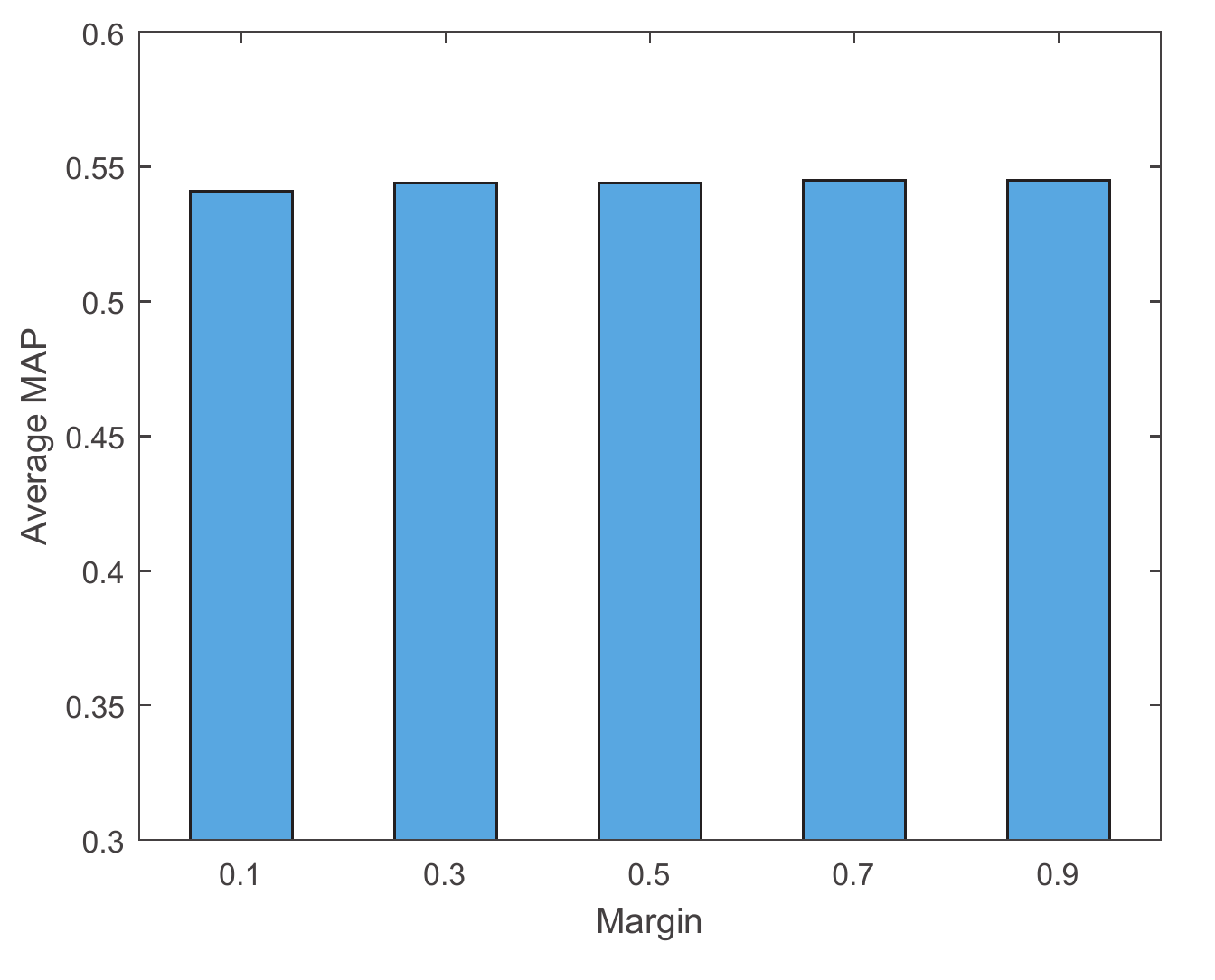}}	
		\setlength{\abovecaptionskip}{0.3cm}
		\caption{Experiments on the influence of the margin parameters, on Wikipedia, Pascal Sentence and XMediaNet datasets. It should be noted that we report the average MAP score of Image$\rightarrow$Text and Text$\rightarrow$Image tasks.
		}
	\label{fig_margin}
	\end{center}
	
\end{figure*}

\subsection{Comparisons with 9 State-of-the-art Methods}

 In this part, we present experimental results on our proposed approach as well as all the compared methods. Tables \ref{table:wiki} to \ref{table:xmn} show the MAP scores of two retrieval tasks and their average scores on 3 datasets, we can see that our proposed approach achieves the best retrieval accuracy compared with 9 state-of-the-art methods. As shown in Table \ref{table:wiki}, our proposed approach has improved the average MAP score from 0.457 to 0.487 on Wikipedia dataset. On one hand, among the compared traditional methods, LGCFL achieves the best retrieval accuracy, which is closer to CMDN based on DNN. On the other hand, the accuracies of 4 DNN-based methods differ greatly, while some of them are outperformed by the traditional methods, such as the average MAP scores of DCCA and Corr-AE are lower than LGCFL and JRL. Besides, results of 2 retrieval tasks as well as their average results on Pascal Sentence and XMediaNet datasets are shown in Tables \ref{table:pas} and \ref{table:xmn}, which have the similar trends as Wikipedia dataset, while our proposed MCSM approach still keeps the best. Some cross-modal retrieval results on XMediaNet dataset of our proposed MCSM approach and the best DNN-based compared method CMDN are shown in Figure \ref{fig_xmn_res}.
 
 Next we give the in-depth analysis on the retrieval results of both our proposed MCSM approach and all compared methods. 
 As shown in Tables \ref{table:wiki} to \ref{table:xmn}, our proposed MCSM approach shows advantage compared with 9 state-of-the-art methods on all 3 datasets. We can also observe that the accuracies of DNN-based methods fail to widen a clear gap with traditional methods. 
 Among the traditional methods, the classical baseline CCA has the worst accuracy for it only models some statistical values, while KCCA extends CCA to achieve better accuracy by the adoption of kernel function with the better ability of modeling nonlinear correlation. CFA has similar results with KCCA, which minimizes the Frobenius norm in the learned common space. JRL and LGCFL are the best two methods among them, while the former utilizes the semi-supervised and sparse regularization, and the latter learns the basis matrices with a local group based priori.

 As for DNN-based methods, DCCA and Corr-AE have close accuracies. DCCA maximizes correlation on the outputs of two separate networks to extend CCA, and Corr-AE not only considers the correlation error but also minimizes the reconstruction error. Deep-SM outperforms DCCA and Corr-AE by fully exploring the strong learning power of convolutional neural network with semantic category information. CMDN achieves the best accuracy among all the compared methods on 2 datasets, because it takes intra-modality and inter-modality correlation into consideration during both separate representation learning and common representation learning stages.
 Compared with all state-of-the-art methods, our proposed MCSM approach has the best accuracy for the fact of following 3 aspects: 
 (1) Independent semantic spaces for different modalities with recurrent attention network to fully exploit the modality-specific fine-grained context information. 
 (2) Attention based joint embedding loss to utilize the imbalanced and complementary relationships between different modalities.
 (3) Adaptive fusion to explore the complementarity between different semantic spaces for cross-modal retrieval.

\subsection{Convergence and Parameter Analysis}

First, we conduct convergence experiments on the relatively large-scale XMediaNet dataset. The curve of downtrend on the loss value is shown in Figure \ref{fig_loss}. We can observe that our proposed approach converges within 15K iterations on the XMediaNet dataset with relatively large scale, which shows its efficiency in training stage. Then, we also conduct parameter experiments on the effect of key parameters, including learning rate and margin parameters $\alpha$ and $\beta$ in equations (\ref{li1}), (\ref{li2}), (\ref{lt1}) and (\ref{lt2}), which are implemented on all the 3 datasets. For the learning rate, we range the value from 1e-2 to 1e-5, and the results are shown in Figure \ref{fig_lr}, from which we can see that our proposed approach achieves the best accuracy at the learning rate of 1e-4 on all the 3 datasets, and the accuracy becomes lower at a higher learning rate. Then, for the margin parameter, it should be noted that we set the margin parameters $\alpha$ and $\beta$ with the same value in all loss functions during each experiment. The value is ranged from 0.1 to 0.9. The results are shown in Figure \ref{fig_margin}, from which we can see that the accuracies are not sensitive to the margin parameters.

 \begin{table*}[htb]
	\caption{Baseline experiments on \textbf{Performance of each semantic space}, where \textbf{MCSM-image} means to retrieve only with the cross-modal similarity generated from image semantic space, while \textbf{MCSM-text} means to only use the cross-modal similarity from text semantic space to retrieve.}
	\begin{center}
		\scalebox{1.0}{
			\begin{tabular}{|c|c|c|c|c|} 
				\hline
				\multirow{2}{*}{Dataset}&
				\multirow{2}{*}{Method} & \multicolumn{3}{c|}{MAP scores} \\
				\cline{3-5}
				& & Image$\rightarrow$Text & Text$\rightarrow$Image & Average \\
				\hline
				
				\multirow{3}{*}{\begin{tabular}{c} Wikipedia dataset \end{tabular}}
				& \textbf{Our MCSM Approach} & \textbf{0.516} & \textbf{0.458} & \textbf{0.487} \\
				
				&   \multirow{1}{*}{\begin{tabular}{c}MCSM-image\end{tabular}} & \multirow{1}{*}{0.448} & \multirow{1}{*}{0.423}& \multirow{1}{*}{0.436}\\
				& \multirow{1}{*}{\begin{tabular}{c}MCSM-text\end{tabular}} & \multirow{1}{*}{0.498} & \multirow{1}{*}{0.438}& \multirow{1}{*}{0.468} \\
				\hline
				
				\multirow{3}{*}{\begin{tabular}{c} Pascal Sentence \\ dataset \end{tabular}} 
				&  \textbf{Our MCSM Approach} & \textbf{0.598} & \textbf{0.598} & \textbf{0.598} \\
				
				&   \multirow{1}{*}{\begin{tabular}{c}MCSM-image\end{tabular}} & \multirow{1}{*}{0.559} & \multirow{1}{*}{0.541}& \multirow{1}{*}{0.550} \\
				& \multirow{1}{*}{\begin{tabular}{c}MCSM-text\end{tabular}} & \multirow{1}{*}{0.500} & \multirow{1}{*}{0.478}& \multirow{1}{*}{0.489} \\
				\hline
				
				\multirow{3}{*}{\begin{tabular}{c} XMediaNet dataset \end{tabular}} 
				&  \textbf{Our MCSM Approach} & \textbf{0.540} & \textbf{0.550} & \textbf{0.545}  \\
				
				&   \multirow{1}{*}{\begin{tabular}{c}MCSM-image\end{tabular}} & \multirow{1}{*}{0.453} & \multirow{1}{*}{0.455}& \multirow{1}{*}{0.454} \\
				& \multirow{1}{*}{\begin{tabular}{c}MCSM-text\end{tabular}} & \multirow{1}{*}{0.447} & \multirow{1}{*}{0.417}& \multirow{1}{*}{0.432} \\
				\hline

			\end{tabular} 
		}
	\end{center}
	%\vspace{-0.5cm}
	\label{table:Baseline1}
\end{table*}

 \begin{figure*}[!t]
	\centering
	\includegraphics[width=1.0\textwidth]{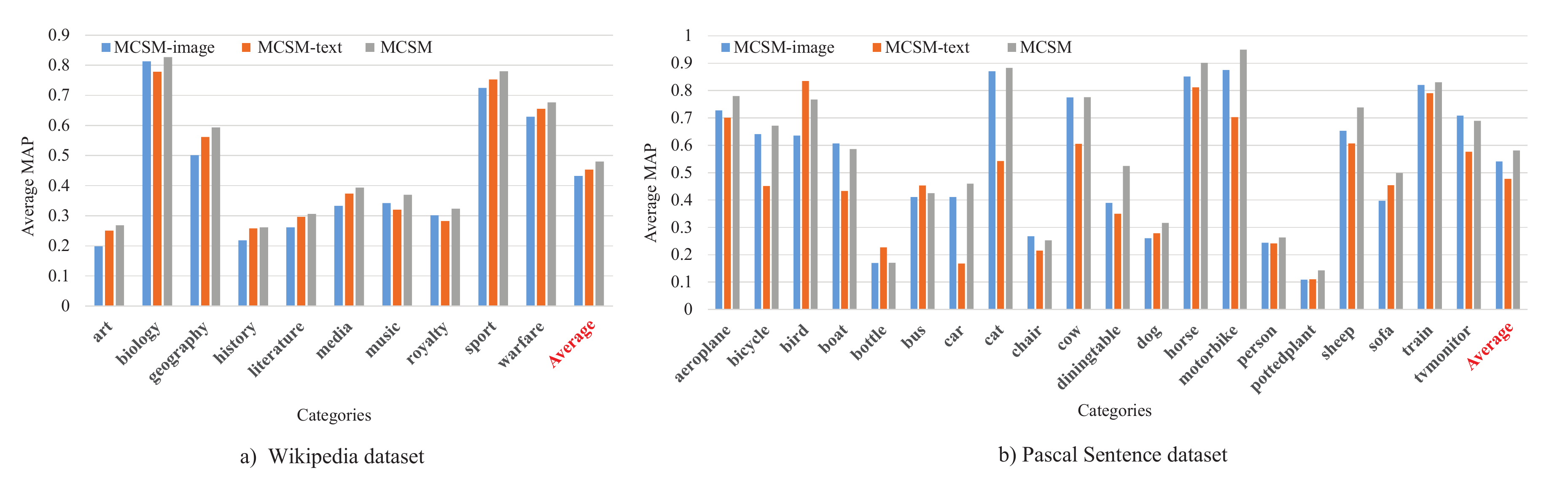}
	\caption{The respective result of each category on our proposed approach, in Wikipedia dataset and Pascal Sentence dataset. We can see that MCSM-text performs better than MCSM-image in most categories of Wikipedia dataset because of the high-level semantic information in text description. While MCSM-image outperforms MCSM-text in Pascal Sentence dataset, because of more useful information in image than only 5 sentences annotated on it. Besides, MCSM achieves the best accuracy in final average results, which indicates the complementarity between two semantic spaces.
	}
	\label{fig_cat}
\end{figure*}

\begin{table*}[htb]
	\caption{Baseline experiments on Performance of \textbf{adaptive fusion} on different semantic spaces, where \textbf{MCSM-LF} means to adopt the cross-modal similarity calculated by late fusion.}
	\begin{center}
		\scalebox{1.0}{
			\begin{tabular}{|c|c|c|c|c|} 
				\hline
				\multirow{2}{*}{Dataset}&
				\multirow{2}{*}{Method} & \multicolumn{3}{c|}{MAP scores} \\
				\cline{3-5}
				& & Image$\rightarrow$Text & Text$\rightarrow$Image & Average \\
				\hline
				
				\multirow{2}{*}{\begin{tabular}{c} Wikipedia dataset \end{tabular}}
				& \textbf{Our MCSM Approach} & \textbf{0.516} & \textbf{0.458} & \textbf{0.487} \\
				
				&   \multirow{1}{*}{\begin{tabular}{c}MCSM-LF\end{tabular}} & \multirow{1}{*}{0.496} & \multirow{1}{*}{0.459}& \multirow{1}{*}{0.478} \\
				\hline
				
				\multirow{2}{*}{\begin{tabular}{c} Pascal Sentence \\ dataset \end{tabular}} 
				&  \textbf{Our MCSM Approach} & \textbf{0.598} & \textbf{0.598} & \textbf{0.598} \\
				
				&   \multirow{1}{*}{\begin{tabular}{c}MCSM-LF\end{tabular}} & \multirow{1}{*}{0.595} & \multirow{1}{*}{0.578}& \multirow{1}{*}{0.587} \\
				\hline
				
				\multirow{2}{*}{\begin{tabular}{c} XMediaNet dataset \end{tabular}} 
				&  \textbf{Our MCSM Approach} & \textbf{0.540} & \textbf{0.550} & \textbf{0.545}  \\
				&   \multirow{1}{*}{\begin{tabular}{c}MCSM-LF\end{tabular}} & \multirow{1}{*}{0.531} & \multirow{1}{*}{0.518}& \multirow{1}{*}{0.525} \\
				\hline

			\end{tabular} 
		}
	\end{center}
	%\vspace{-0.5cm}
	\label{table:Baseline2}
\end{table*}

\subsection{Baseline Comparisons}

In this part, we conduct two baseline experiments as follows, to verify the separate contribution of each component in our proposed MCSM approach. Tables \ref{table:Baseline1} and \ref{table:Baseline2} show the accuracy of our proposed MCSM approach as well as the baseline approaches on the following two aspects.

\subsubsection{Performance of each semantic space}

As shown in Table \ref{table:Baseline1}, MCSM-image means to perform cross-modal retrieval only by the cross-modal similarity $sim_i$ in equation (\ref{sim_i}) generated from image semantic space, while MCSM-text means to use the cross-modal similarity $sim_t$ only in equation (\ref{sim_t}) from text semantic space. We can observe that MCSM-text has better accuracy than MCSM-image on Wikipedia dataset, while on Pascal Sentence dataset and XMediaNet dataset, MCSM-image outperforms MCSM-text in the average MAP score. 

This is because of the imbalanced and complementary relationships between different modalities that contain unequal amount of information. The respective result of each category is reported in Figure \ref{fig_cat}, taking Wikipedia and Pascal Sentence datasets as examples. Specifically, in Wikipedia dataset, the categories mostly lie in high-level semantics, such as history or literature, where the textual description contains more background information that cannot be presented by its corresponding image. As for the other two datasets, their categories mostly are specific objects, including animal such as elephant, or artifact such as airplane. Their corresponding textual descriptions are relatively simple, such as only 5 sentences to describe each image in Pascal Sentence dataset. Under this situation, the visual description contains more useful information than its corresponding text, which leads to the higher accuracy of MCSM-image. Besides, the final accuracy of MCSM stably outperforms MCSM-image and MCSM-text, which indicates that there exists reasonable complementarity between the two semantic spaces. Thus, from the above results, the motivation of this paper is fully verified.

\subsubsection{Performance of adaptive fusion on different semantic spaces}

We also present the baseline experiments to verify the effectiveness of adaptive fusion on different semantic spaces. We compare our proposed adaptive fusion strategy with late fusion (MCSM-LF), which means to directly average the two kinds of cross-modal similarities generated from different semantic spaces by the following equation.
\begin{align}
sim_{lf}(i_p,t_p)=\frac{1}{2}(sim_i(i_p,t_p)+sim_t(i_p,t_p)).
\end{align}
The results on 3 datasets are shown in Table \ref{table:Baseline2}, where MCSM-LF means the accuracy calculated by late fusion. We can observe that adaptive fusion can further exploit the rich complementary information between the two semantic spaces to boost the accuracy of cross-modal retrieval.

From the above baseline results, the separate contribution of each component in our proposed MCSM approach can be verified. First, the imbalanced and complementary relationships between different modalities are fully exploited in different semantic spaces. Second, the complementarity between different semantic spaces is fully captured by the adaptive fusion.

\section{Conclusion}
In this paper, we have proposed a modality-specific cross-modal similarity measurement approach to construct independent semantic spaces for different modalities. First, recurrent attention network with attention based joint embedding loss is adopted to fully model the modality-specific characteristics within each modality, and utilize the imbalanced and complementary relationships between different modalities during correlation learning. Second, the end-to-end frameworks are implemented in different semantic spaces to directly generate the cross-modality similarity, integrating both common representation learning and distance metric learning to benefit each other. Third, the adaptive fusion is adopted to explore the complementarity between different semantic space. Experiments on 3 cross-modal datasets, including widely-used Wikipedia and Pascal Sentence datasets as well as our constructed large-scale XMediaNet dataset, verify the effectiveness of our proposed approach compared with 9 state-of-the-art methods.

As for the future work, we attempt to extend the current framework to other modalities such as video, audio and so on, for exploring the imbalanced and complementary relationships across the data of multiple modalities. Besides, we attempt to transfer knowledge from external knowledge base to further boost the performance of cross-modal retrieval.

\ifCLASSOPTIONcaptionsoff
  \newpage
\fi



%\bibliographystyle{IEEEtran}
\bibliographystyle{IEEEtran}
\bibliography{ijcai16}

\begin{thebibliography}{10}
\providecommand{\url}[1]{#1}
\csname url@samestyle\endcsname
\providecommand{\newblock}{\relax}
\providecommand{\bibinfo}[2]{#2}
\providecommand{\BIBentrySTDinterwordspacing}{\spaceskip=0pt\relax}
\providecommand{\BIBentryALTinterwordstretchfactor}{4}
\providecommand{\BIBentryALTinterwordspacing}{\spaceskip=\fontdimen2\font plus
\BIBentryALTinterwordstretchfactor\fontdimen3\font minus
  \fontdimen4\font\relax}
\providecommand{\BIBforeignlanguage}[2]{{%
\expandafter\ifx\csname l@#1\endcsname\relax
\typeout{** WARNING: IEEEtran.bst: No hyphenation pattern has been}%
\typeout{** loaded for the language `#1'. Using the pattern for}%
\typeout{** the default language instead.}%
\else
\language=\csname l@#1\endcsname
\fi
#2}}
\providecommand{\BIBdecl}{\relax}
\BIBdecl

\bibitem{DBLP:journals/tip/TangLWZ15}
J.~Tang, Z.~Li, M.~Wang, and R.~Zhao, ``Neighborhood discriminant hashing for
  large-scale image retrieval,'' \emph{IEEE Transactions on Image Processing
  (TIP)}, vol.~24, no.~9, pp. 2827--2840, 2015.

\bibitem{PengCSVT06ClipRetrieval}
Y.~Peng and C.-W. Ngo, ``Clip-based similarity measure for query-dependent clip
  retrieval and video summarization,'' \emph{IEEE Transactions on Circuits and
  Systems for Video Technology (TCSVT)}, vol.~16, no.~5, pp. 612--627, 2006.

\bibitem{peng2017overview}
Y.~Peng, X.~Huang, and Y.~Zhao, ``An overview of cross-media retrieval:
  Concepts, methodologies, benchmarks and challenges,'' \emph{IEEE Transactions
  on Circuits and Systems for Video Technology (TCSVT)}, 2017.

\bibitem{RasiwasiaMM10SemanticCCA}
N.~Rasiwasia, J.~Costa~Pereira, E.~Coviello, G.~Doyle, G.~R. Lanckriet,
  R.~Levy, and N.~Vasconcelos, ``A new approach to cross-modal multimedia
  retrieval,'' in \emph{ACM International Conference on Multimedia (ACM-MM)},
  2010, pp. 251--260.

\bibitem{DBLP:journals/tip/XuSYSL17}
X.~Xu, F.~Shen, Y.~Yang, H.~T. Shen, and X.~Li, ``Learning discriminative
  binary codes for large-scale cross-modal retrieval,'' \emph{IEEE Transactions
  on Image Processing (TIP)}, vol.~26, no.~5, pp. 2494--2507, 2017.

\bibitem{DBLP:journals/tip/WangGWHY16}
D.~Wang, X.~Gao, X.~Wang, L.~He, and B.~Yuan, ``Multimodal discriminative
  binary embedding for large-scale cross-modal retrieval,'' \emph{IEEE
  Transactions on Image Processing (TIP)}, vol.~25, no.~10, pp. 4540--4554,
  2016.

\bibitem{DBLP:journals/tip/HeZWJY15}
R.~He, M.~Zhang, L.~Wang, Y.~Ji, and Q.~Yin, ``Cross-modal subspace learning
  via pairwise constraints,'' \emph{IEEE Transactions on Image Processing
  (TIP)}, vol.~24, no.~12, pp. 5543--5556, 2015.

\bibitem{DBLP:journals/ijcv/GongKIL14}
Y.~Gong, Q.~Ke, M.~Isard, and S.~Lazebnik, ``A multi-view embedding space for
  modeling internet images, tags, and their semantics,'' \emph{International
  Journal of Computer Vision (IJCV)}, vol. 106, no.~2, pp. 210--233, 2014.

\bibitem{ZhaiTCSVT2014JRL}
X.~Zhai, Y.~Peng, and J.~Xiao, ``Learning cross-media joint representation with
  sparse and semi-supervised regularization,'' \emph{IEEE Transactions on
  Circuits and Systems for Video Technology (TCSVT)}, vol.~24, pp. 965--978,
  2014.

\bibitem{DBLP:journals/pami/WangHWWT16}
K.~Wang, R.~He, L.~Wang, W.~Wang, and T.~Tan, ``Joint feature selection and
  subspace learning for cross-modal retrieval,'' \emph{IEEE Transactions on
  Pattern Analysis and Machine Intelligence (TPAMI)}, vol.~38, no.~10, pp.
  2010--2023, 2016.

\bibitem{DBLP:journals/tip/WuJLTLZZ15}
F.~Wu, X.~Jiang, X.~Li, S.~Tang, W.~Lu, Z.~Zhang, and Y.~Zhuang, ``Cross-modal
  learning to rank via latent joint representation,'' \emph{IEEE Transactions
  on Image Processing (TIP)}, vol.~24, no.~5, pp. 1497--1509, 2015.

\bibitem{DBLP:journals/tip/WuLSYZRZ16}
F.~Wu, X.~Lu, J.~Song, S.~Yan, Z.~M. Zhang, Y.~Rui, and Y.~Zhuang, ``Learning
  of multimodal representations with random walks on the click graph,''
  \emph{IEEE Transactions on Image Processing (TIP)}, vol.~25, no.~2, pp.
  630--642, 2016.

\bibitem{DBLP:conf/ijcai/PengHQ16}
Y.~Peng, X.~Huang, and J.~Qi, ``Cross-media shared representation by
  hierarchical learning with multiple deep networks,'' in \emph{International
  Joint Conference on Artificial Intelligence (IJCAI)}, 2016, pp. 3846--3853.

\bibitem{feng12014cross}
F.~Feng, X.~Wang, and R.~Li, ``Cross-modal retrieval with correspondence
  autoencoder,'' in \emph{ACM International Conference on Multimedia (ACM-MM)},
  2014, pp. 7--16.

\bibitem{DBLP:conf/cvpr/YanM15}
F.~Yan and K.~Mikolajczyk, ``Deep correlation for matching images and text,''
  in \emph{Conference on Computer Vision and Pattern Recognition (CVPR)}, 2015,
  pp. 3441--3450.

\bibitem{DBLP:conf/icml/XuBKCCSZB15}
K.~Xu, J.~Ba, R.~Kiros, K.~Cho, A.~C. Courville, R.~Salakhutdinov, R.~S. Zemel,
  and Y.~Bengio, ``Show, attend and tell: Neural image caption generation with
  visual attention,'' in \emph{International Conference on Machine Learning
  (ICML)}, 2015, pp. 2048--2057.

\bibitem{DBLP:conf/nips/LuYBP16}
J.~Lu, J.~Yang, D.~Batra, and D.~Parikh, ``Hierarchical question-image
  co-attention for visual question answering,'' in \emph{Conference on Neural
  Information Processing Systems (NIPS)}, 2016, pp. 289--297.

\bibitem{HotelingBiometrika36RelationBetweenTwoVariates}
H.~Hotelling, ``Relations between two sets of variates,'' \emph{Biometrika},
  pp. 321--377, 1936.

\bibitem{DBLP:journals/pami/PereiraCDRLLV14}
J.~C. Pereira, E.~Coviello, G.~Doyle, N.~Rasiwasia, G.~R.~G. Lanckriet,
  R.~Levy, and N.~Vasconcelos, ``On the role of correlation and abstraction in
  cross-modal multimedia retrieval,'' \emph{IEEE Transactions on Pattern
  Analysis and Machine Intelligence (TPAMI)}, vol.~36, no.~3, pp. 521--535,
  2014.

\bibitem{DBLP:conf/iccv/RanjanRJ15}
V.~Ranjan, N.~Rasiwasia, and C.~V. Jawahar, ``Multi-label cross-modal
  retrieval,'' in \emph{IEEE International Conference on Computer Vision
  (ICCV)}, 2015, pp. 4094--4102.

\bibitem{LiMM03CFA}
D.~Li, N.~Dimitrova, M.~Li, and I.~K. Sethi, ``Multimedia content processing
  through cross-modal association,'' in \emph{ACM International Conference on
  Multimedia (ACM-MM)}, 2003, pp. 604--611.

\bibitem{DBLP:conf/colt/BelkinMN04}
M.~Belkin, I.~Matveeva, and P.~Niyogi, ``Regularization and semi-supervised
  learning on large graphs,'' in \emph{Annual Conference on Learning Theory
  (COLT)}, 2004, pp. 624--638.

\bibitem{ZhaiAAAI2013JGRHML}
X.~Zhai, Y.~Peng, and J.~Xiao, ``Heterogeneous metric learning with joint graph
  regularization for cross-media retrieval,'' in \emph{AAAI Conference on
  Artificial Intelligence (AAAI)}, 2013, pp. 1198--1204.

\bibitem{DBLP:journals/tcsv/PengZZH16}
Y.~Peng, X.~Zhai, Y.~Zhao, and X.~Huang, ``Semi-supervised cross-media feature
  learning with unified patch graph regularization,'' \emph{IEEE Transactions
  on Circuits and Systems for Video Technology (TCSVT)}, vol.~26, no.~3, pp.
  583--596, 2016.

\bibitem{DBLP:conf/nips/KrizhevskySH12}
A.~Krizhevsky, I.~Sutskever, and G.~E. Hinton, ``Imagenet classification with
  deep convolutional neural networks,'' in \emph{Conference on Neural
  Information Processing Systems (NIPS)}, 2012, pp. 1106--1114.

\bibitem{DBLP:conf/mm/WuJWYX16}
Z.~Wu, Y.~Jiang, X.~Wang, H.~Ye, and X.~Xue, ``Multi-stream multi-class fusion
  of deep networks for video classification,'' in \emph{ACM International
  Conference on Multimedia (ACM-MM)}, 2016, pp. 791--800.

\bibitem{DBLP:conf/cvpr/HeZRS16}
K.~He, X.~Zhang, S.~Ren, and J.~Sun, ``Deep residual learning for image
  recognition,'' in \emph{Conference on Computer Vision and Pattern Recognition
  (CVPR)}, 2016, pp. 770--778.

\bibitem{ngiam32011multimodal}
J.~Ngiam, A.~Khosla, M.~Kim, J.~Nam, H.~Lee, and A.~Y. Ng, ``Multimodal deep
  learning,'' in \emph{International Conference on Machine Learning (ICML)},
  2011, pp. 689--696.

\bibitem{srivastava2012learning}
N.~Srivastava and R.~Salakhutdinov, ``Learning representations for multimodal
  data with deep belief nets,'' in \emph{International Conference on Machine
  Learning (ICML) Workshop}, 2012.

\bibitem{DBLP:conf/icml/AndrewABL13}
G.~Andrew, R.~Arora, J.~A. Bilmes, and K.~Livescu, ``Deep canonical correlation
  analysis,'' in \emph{International Conference on Machine Learning (ICML)},
  2013, pp. 1247--1255.

\bibitem{DBLP:journals/tcyb/WeiZLWLZY17}
Y.~Wei, Y.~Zhao, C.~Lu, S.~Wei, L.~Liu, Z.~Zhu, and S.~Yan, ``Cross-modal
  retrieval with {CNN} visual features: {A} new baseline,'' \emph{IEEE
  Transactions on Cybernetics (TCYB)}, vol.~47, no.~2, pp. 449--460, 2017.

\bibitem{DBLP:journals/tmm/HeXKWP16}
Y.~He, S.~Xiang, C.~Kang, J.~Wang, and C.~Pan, ``Cross-modal retrieval via deep
  and bidirectional representation learning,'' \emph{IEEE Transactions on
  Multimedia (TMM)}, vol.~18, no.~7, pp. 1363--1377, 2016.

\bibitem{DBLP:conf/nips/MnihHGK14}
V.~Mnih, N.~Heess, A.~Graves, and K.~Kavukcuoglu, ``Recurrent models of visual
  attention,'' in \emph{Conference on Neural Information Processing Systems
  (NIPS)}, 2014, pp. 2204--2212.

\bibitem{DBLP:conf/icml/GregorDGRW15}
K.~Gregor, I.~Danihelka, A.~Graves, D.~J. Rezende, and D.~Wierstra, ``{DRAW:}
  {A} recurrent neural network for image generation,'' in \emph{International
  Conference on Machine Learning (ICML)}, 2015, pp. 1462--1471.

\bibitem{DBLP:conf/cvpr/YangHGDS16}
Z.~Yang, X.~He, J.~Gao, L.~Deng, and A.~J. Smola, ``Stacked attention networks
  for image question answering,'' in \emph{Conference on Computer Vision and
  Pattern Recognition (CVPR)}, 2016, pp. 21--29.

\bibitem{DBLP:journals/corr/RocktaschelGHKB15}
T.~Rockt{\"{a}}schel, E.~Grefenstette, K.~M. Hermann, T.~Kocisk{\'{y}}, and
  P.~Blunsom, ``Reasoning about entailment with neural attention,'' in
  \emph{International Conference on Learning Representations (ICLR)}, 2016.

\bibitem{DBLP:conf/nips/HermannKGEKSB15}
K.~M. Hermann, T.~Kocisk{\'{y}}, E.~Grefenstette, L.~Espeholt, W.~Kay,
  M.~Suleyman, and P.~Blunsom, ``Teaching machines to read and comprehend,'' in
  \emph{Conference on Neural Information Processing Systems (NIPS)}, 2015, pp.
  1693--1701.

\bibitem{DBLP:conf/emnlp/RushCW15}
A.~M. Rush, S.~Chopra, and J.~Weston, ``A neural attention model for
  abstractive sentence summarization,'' in \emph{Conference on Empirical
  Methods in Natural Language Processing (EMNLP)}, 2015, pp. 379--389.

\bibitem{DBLP:journals/corr/SimonyanZ14a}
K.~Simonyan and A.~Zisserman, ``Very deep convolutional networks for
  large-scale image recognition,'' in \emph{International Conference on
  Learning Representations (ICLR)}, 2014.

\bibitem{DBLP:journals/corr/LinPWZ17}
Y.~Lin, Z.~Pang, D.~Wang, and Y.~Zhuang, ``Task-driven visual saliency and
  attention-based visual question answering,'' \emph{arXiv preprint
  arXiv:1702.06700}, 2017.

\bibitem{DBLP:journals/neco/HochreiterS97}
S.~Hochreiter and J.~Schmidhuber, ``Long short-term memory,'' \emph{Neural
  Computation}, vol.~9, no.~8, pp. 1735--1780, 1997.

\bibitem{DBLP:conf/nips/MikolovSCCD13}
T.~Mikolov, I.~Sutskever, K.~Chen, G.~S. Corrado, and J.~Dean, ``Distributed
  representations of words and phrases and their compositionality,'' in
  \emph{Conference on Neural Information Processing Systems (NIPS)}, 2013, pp.
  3111--3119.

\bibitem{DBLP:conf/emnlp/Kim14}
Y.~Kim, ``Convolutional neural networks for sentence classification,'' in
  \emph{Conference on Empirical Methods in Natural Language Processing
  (EMNLP)}, 2014, pp. 1746--1751.

\bibitem{DBLP:journals/corr/KongDDZT16}
G.~Kong, L.~Dong, W.~Dong, L.~Zheng, and Q.~Tian, ``Coarse2fine: Two-layer
  fusion for image retrieval,'' \emph{arXiv preprint arXiv:1607.00719}, 2016.

\bibitem{rashtchian2010collecting}
C.~Rashtchian, P.~Young, M.~Hodosh, and J.~Hockenmaier, ``Collecting image
  annotations using amazon's mechanical turk,'' in \emph{NAACL HLT 2010
  Workshop on Creating Speech and Language Data with Amazon's Mechanical Turk},
  2010, pp. 139--147.

\bibitem{DBLP:journals/neco/HardoonSS04}
D.~R. Hardoon, S.~Szedm{\'{a}}k, and J.~Shawe{-}Taylor, ``Canonical correlation
  analysis: An overview with application to learning methods,'' \emph{Neural
  Computation}, vol.~16, no.~12, pp. 2639--2664, 2004.

\bibitem{DBLP:journals/tmm/KangXLXP15}
C.~Kang, S.~Xiang, S.~Liao, C.~Xu, and C.~Pan, ``Learning consistent feature
  representation for cross-modal multimedia retrieval,'' \emph{IEEE
  Transactions on Multimedia (TMM)}, vol.~17, no.~3, pp. 370--381, 2015.

\end{thebibliography}

%\end{IEEEbiography}
%% insert where needed to balance the two columns on the last page with
%%\newpage


%\vfill


\end{document}